\documentclass[11pt]{article}
\usepackage[numbers]{natbib}
\usepackage{graphicx, float}
\usepackage{amsmath, amsfonts, amsthm, amssymb, bm, mathtools}
\usepackage{geometry}
\usepackage{tikz}
\usetikzlibrary{patterns}
\usepackage{xcolor}
\usepackage{url}
\usepackage[colorlinks=true]{hyperref} \hypersetup{urlcolor=blue, citecolor=red}
\usepackage{cleveref}
\usepackage{todonotes}

\newtheorem{theorem}{Theorem}[section]

\newtheorem{assumption}[theorem]{Assumption}

\newtheorem{remark}{Remark}
\newtheorem{example}{Example}

\DeclarePairedDelimiter\iprod{\langle}{\rangle}
\DeclarePairedDelimiter\bnorm{\lVert}{\rVert}

\newcommand{\bsc}{{\bm c}} 
\newcommand{\bse}{{\bm e}} 
\newcommand{\bsv}{{\bm v}}
\newcommand{\bsx}{{\bm x}}

\newcommand{\bsmu}{{\bm \mu}} \newcommand{\bsnu}{{\bm \nu}}
 
\newcommand{\bsxi}{{\bm \xi}}

\newcommand{\C}{{\mathbb C}}
\newcommand{\E}{{\mathbb E}}
\newcommand{\N}{{\mathbb N}}
\newcommand{\R}{{\mathbb R}}
\newcommand{\bbP}{{\mathbb P}}
\newcommand{\bbT}{{\mathbb T}}

\newcommand{\cO}{\mathcal{O}}

\newcommand{\dd}{\;\mathrm{d}}

\newcommand{\set}[2]{\{#1\,:\,#2\}}
\newcommand{\norm}[2][]{\| #2 \|_{{#1}}}
\newcommand{\normOut}[1]{\| #1 \|_{{\CY}}}

\newcommand{\CEinv}[1][]{\ifx\relax#1\relax \mathcal{E}^{\dagger} \else \mathcal{E}^{-1}_{\dagger} \fi}
\newcommand{\CDinv}[1][]{\ifx\relax#1\relax \mathcal{D}^{\dagger} \else \mathcal{D}^{\dagger}_{#1} \fi}

\newcommand{\LTwoRBNO}{\tilde{\CG}_{L^2_\mu-{\text{RB-NO}}}}
\newcommand{\HOneRBNO}{\tilde{\CG}_{H^1_\mu-{\text{RB-NO}}}}
\newcommand{\FNO}{\tilde{\CG}_{\text{FNO}}}
\newcommand{\Smolyak}{\tilde{\CG}_{\text{RB-SG}}}
\newcommand{\TT}{\tilde{\CG}_{\text{RB-TT}}}

\providecommand{\BA}{{\boldsymbol{A}}}
\providecommand{\BK}{{\boldsymbol{K}}}
\providecommand{\Bb}{{\boldsymbol{b}}}

\providecommand{\Bu}{{\boldsymbol{u}}}
\providecommand{\Bv}{{\boldsymbol{v}}}

\providecommand{\CL}{{\mathcal{L}}}
\newcommand{\Blambda}    {{\boldsymbol{\lambda}}}

\newcommand{\CX}{\mathcal{X}}
\newcommand{\CY}{\mathcal{Y}}
\newcommand{\CG}{\mathcal{G}}
\newcommand{\CR}{\mathcal{R}}
\newcommand{\CF}{\mathcal{F}}
\newcommand{\CE}{\mathcal{E}}
\newcommand{\CD}{\mathcal{D}}

\title{Performance of Neural and Polynomial Operator Surrogates}
\author{Josephine Westermann\thanks{Heidelberg University}
\and Benno Huber\footnotemark[1]
\and Thomas O'Leary-Roseberry\thanks{The Ohio State University}
\and Jakob Zech\footnotemark[1]}
\date{\today}

\begin{document}

\maketitle

\begin{abstract}
We consider the problem of constructing surrogate operators for parameter-to-solution maps arising from parametric partial differential equations, where repeated forward model evaluations are computationally expensive. We present a systematic empirical comparison of neural operator surrogates, including a reduced-basis neural operator trained with $L^2_\mu$ and $H^1_\mu$ objectives and the Fourier neural operator, against polynomial surrogate methods, specifically a reduced-basis sparse-grid surrogate and a reduced-basis tensor-train surrogate.
All methods are evaluated on a linear parametric diffusion problem and a nonlinear parametric hyperelasticity problem, using input fields with algebraically decaying spectral coefficients at varying rates of decay $s$. To enable fair comparisons, we analyze ensembles of surrogate models generated by varying hyperparameters and compare the resulting Pareto frontiers of cost versus approximation accuracy, decomposing cost into contributions from data generation, setup, and evaluation.
Our results show that no single method is universally superior.
Polynomial surrogates achieve substantially better data efficiency for smooth input fields ($s \geq 2$), with convergence rates for the sparse-grid surrogate in agreement with theoretical predictions. For rough inputs ($s \leq 1$), the Fourier neural operator displays the fastest convergence rates. Derivative-informed training consistently improves data efficiency over standard $L^2_\mu$ training, providing a competitive alternative for rough inputs in the low-data regime when Jacobian information is available at reasonable cost.
These findings highlight the importance of matching the surrogate methodology to the regularity of the problem as well as accuracy demands and computational constraints of the application.
\end{abstract}



\section{Introduction}

Consider a (generally nonlinear) operator $\mathcal G : \mathcal X \to \mathcal Y$ between separable Hilbert spaces. Such operators arise ubiquitously in science and engineering as parameter-to-observable maps defined by underlying physical models, most commonly partial differential equations (PDEs). In these settings, $\mathcal G$ is typically defined implicitly by the solution of a forward problem
\[
\mathcal G(x) = y
\quad \text{such that} \quad
R(y,x) = 0 \in \mathcal Y',
\]
where $R$ denotes the residual of the governing model and takes values in the topological dual $\mathcal Y'$ of the output space.

We assume that $\mathcal G$ can be evaluated pointwise, although such evaluations may be computationally expensive and the operator need not be linear. Typical examples include parameter-to-solution maps associated with partial differential equations (PDEs), as well as associated parameter-to-observable maps. Many problems in modern applied mathematics require a large number of evaluations of $\mathcal G$, e.g.\ stochastic simulation, inverse problems, and PDE-constrained optimization. When the cost of these evaluations becomes prohibitive, a natural alternative is to construct a surrogate operator $\widetilde{\mathcal G}$ from a limited number of evaluations. The surrogate is designed to be inexpensive to evaluate while accurately approximating $\mathcal G$ over a prescribed input distribution $\mu$, for example in the sense that
\[
\mathbb{E}_{x\sim \mu}\left[\|\mathcal{G}(x) - \mathcal{G}_\theta(x)\|^2_\mathcal{Y}\right] = \int_{\mathcal X} \bigl\|\mathcal G(x) - \widetilde{\mathcal G}(x)\bigr\|_{\mathcal Y}^2 \, \mathrm d\mu(x) < \varepsilon^2 ,
\]
for an appropriately small $\varepsilon>0$.

Constructing accurate operator surrogates that are computationally inexpensive to evaluate has become an active area of research in recent years.
A wide range of methods, often referred to as \emph{operator learning}, have been developed to approximate high-dimensional maps arising from PDE solution operators and other parametric models. In this work, we investigate the fundamental tradeoffs between different operator learning architectures and formulations, with a particular focus on their accuracy, efficiency, and suitability for high-dimensional PDE-induced maps.

\subsection{Operator learning architectures and formulations}

\paragraph{Reduced basis methods}

A common strategy for reducing the computational cost of evaluating high-dimensional operator maps is to exploit low-dimensional structure through dimension reduction. In the context of PDE-based models, this typically takes the form of identifying low-dimensional linear subspaces of the discretized state space in which the solution manifold is well approximated. Classical reduced basis methods construct such subspaces using techniques such as proper orthogonal decomposition (POD) \cite{hinze2005proper,volkwein2013proper}, principal component analysis (PCA) \cite{ramsay1997functional,bhattacharya2021model}, and derivative-informed subspaces \cite{zahm2020gradient,o2022derivative}, yielding representations in which the input--output map can be expressed in terms of a small number of basis coefficients that is independent of the underlying discretization. These bases can often be computed efficiently via generalized eigenvalue problems or related spectral decompositions in an offline stage, making them attractive for large-scale models \cite{saibaba2016randomized}. Once a reduced basis has been constructed, a wide variety of operator learning architectures can be formulated on the reduced coordinates, significantly lowering the cost of both training and evaluation. Although nonlinear dimension reduction techniques from machine learning offer increased expressivity, they typically sacrifice discretization-independent structure, stability, and interpretability, and therefore are not considered in this work.

\paragraph{Polynomial operator surrogates}

Polynomial surrogates, including in particular polynomial chaos expansions (PCE) \cite{wiener1938homogeneous,xiu2002wiener,xiu2010numerical} and sparse (-grid) polynomial approximations \cite{nobile2008sparsegridcollocation, nobile2008anisotropicsparsegridcollocation, cohen2011analytic}, provide a framework for approximating parameter-to-solution maps by global expansions in multivariate polynomials. These methods offer strong theoretical guarantees and can achieve rapid convergence for operators that depend smoothly on the parameters, even when the dependence is highly nonlinear. As is well-known, for high-dimensional problems, their effectiveness typically relies on the presence of a low-dimensional or strongly anisotropic parametric structure that can be exploited by the scheme \cite{cohen2010convergence,chkifa2015breaking}.

\paragraph{Tensor-train surrogates}

Low-rank tensor surrogate models provide an alternative to \emph{sparse} polynomial methods; they represent the parameter-to-solution map on a \emph{full} tensor-product collocation grid, and achieve compression by determining an (approximate) low-rank tensor decomposition of the resulting multidimensional array of values \cite{KoldaBader2009}. Among such decompositions, the tensor-train (TT) format \cite{osedelets11} is particularly attractive due to its simple, sequential structure and favorable scaling with dimension. Typically, a polynomial-based interpolation is used to evaluate the surrogate in between grid points. This representation enables the development of numerical algorithms that operate directly in compressed form, such as the ALS-cross method \cite{dolgov18}, which constructs TT surrogates of parametric PDE solution operators from a limited number of model evaluations. In recent work, tensor-train surrogates have been developed to construct local high-order Taylor approximations of input-output maps \cite{alger2026taylor}.

\paragraph{Neural operators}

In recent years, neural network–based surrogate models have emerged as a powerful approach for learning input–output maps defined by parametric PDE solution operators. A particularly important class of methods, often referred to as \emph{neural operators}, is designed to approximate mappings between infinite-dimensional function spaces rather than fixed finite-dimensional vectors \cite{kovachki2023neural}. These architectures typically incorporate discretization-invariant design principles, such as representing inputs and outputs in spectral or reduced bases, or learning approximations to integral kernel operators in compressed representations.

Prominent examples include the Fourier neural operator (FNO) \cite{li2021fourier}, variants of the DeepONet architecture \cite{lu2022comprehensive}, graph-based methods \cite{li2020neural, li2020multipole}, transformer architectures \cite{calvello2025continuum}, and approaches that combine neural networks with reduced-basis representations, such as PCANet \cite{hesthaven2018non,bhattacharya2021model}. These methods possess several attractive features, including strong approximation properties as well as flexibility in implementation and training; especially universal approximation can usually be established under very mild assumptions, and holds for essentially all currently popular methods, e.g.\ \cite{lu2021learning,kovachki2021universal,lanthaler2025nonlocality,SSZ23_3044}. Quantitative rates are also available in certain settings \cite{kovachki2021universal, lanthaler2022error, lanthaler2023operator, herrmann2024neural}.

Neural operator architectures and related approaches, including graph-based approaches and spectral approaches, have demonstrated impressive predictive capability in complex, high-dimensional applications, including challenging problems in scientific computing such as numerical weather prediction \cite{bi2022pangu,lam2023learning,pathak2022fourcastnet,pathak2026learning,kossaifi2026demystifying}.

\paragraph{Training formulations}

In practice, neural operators are typically constructed by empirical risk minimization over sampled input–output pairs, for example by approximating an $L^2_\mu$ objective of the form
\begin{equation*}
    \min_\theta \; \mathbb{E}_{x \sim \mu}
    \bigl[ \| \mathcal{G}(x) - \mathcal{G}_\theta(x)\|_{\mathcal{Y}}^2 \bigr],
\end{equation*}
or, in physics-informed settings, by minimizing a residual-based objective \cite{li2024physics,qiu2025variationally},
\begin{equation*}
    \min_\theta \; \mathbb{E}_{x \sim \mu}
    \bigl[ \| R(\mathcal{G}_\theta(x),x) \|_{\mathcal{Y}'}^2 \bigr].
\end{equation*}
In addition, we consider derivative-informed training strategies \cite{OLEARYROSEBERRY2024}, in which the neural operator is trained to approximately minimize an $H^1_\mu$-type objective:
\begin{equation*}
    \min_\theta \; \mathbb{E}_{x \sim \mu}
    \Bigl[ \| \mathcal{G}(x) - \mathcal{G}_\theta(x)\|_{\mathcal{Y}}^2
    + \|D\mathcal{G}(x) - D\mathcal{G}_\theta(x)\|^2 \Bigr],
  \end{equation*}
where $D\mathcal{G}(x) \in \mathcal{L}(\mathcal{X},\mathcal{Y})$ denotes the Fr\'echet derivative, and the norm on the derivative term must be chosen carefully, to ensure computational efficiency, see \Cref{sec:dertraining} ahead.

In practice, derivative-informed training can be made computationally feasible by exploiting low-dimensional structure in the derivatives, in the neural operator architecture, or both—for example, through dimension-reduction strategies. Incorporating derivative information can improve approximation accuracy in $L^2_\mu$ and enhance performance in downstream tasks such as optimization and inverse problems~~\cite{yao2025derivative,luo2025efficient,gong2026shape,cao2026lazydino}.

\subsection{Related work}

\paragraph{Neural operator cost-accuracy comparisons.} Several recent studies have conducted systematic comparisons of neural operator architectures for learning solution operators of PDEs. One line of work performs large-scale benchmarking across multiple PDE problems using standardized datasets and training protocols, comparing architectures such as DeepONet and the Fourier neural operator primarily in terms of prediction error and robustness across geometries and noise levels \cite{lu2022comprehensive}. Another benchmarking study evaluates a range of neural operator models through the lens of the cost--accuracy trade-off, measuring the computational cost required to achieve a given approximation error across PDE benchmarks \cite{de2022cost}. No single neural operator architecture consistently dominates across benchmarks. In particular, the Fourier neural operator is notably parameter efficient but is most naturally suited to structured domains, whereas other architectures offer greater flexibility for handling more general geometries.

\paragraph{Benchmarking practices in operator learning.}
In \cite{mcgreivy2024weak} the authors examine benchmarking practices in operator learning and emphasize the importance of strong and well-tuned baselines. Their analysis shows that classical numerical methods often substantially outperform modern operator-learning architectures, motivating careful evaluation across a broad range of surrogate approaches.

\subsection{Contributions}

We present a systematic comparison of neural operator surrogates and polynomial surrogate models for approximating parameter-to-solution maps of PDEs. In contrast to the aforementioned benchmarking studies, which focus primarily on comparisons among neural operator architectures, we include surrogate models with well-established approximation theory, including sparse polynomial expansions and tensor-train representations, which under suitable regularity assumptions can achieve dimension-independent convergence rates.

We evaluate surrogate models across a range of operator smoothness regimes, highlighting how approximation performance depends on the regularity of the underlying parameter-to-solution map. To enable fair comparison, we evaluate ensembles of surrogate models obtained by varying hyperparameters and compare the resulting Pareto frontiers of computational cost versus approximation accuracy. We empirically investigate convergence with respect to the number of training samples (operator evaluations) and quantify the computational cost of surrogate evaluation, providing a comprehensive picture of the trade-offs between data efficiency, computational efficiency, and approximation accuracy.

\subsection{Outline}

In \Cref{sec:enc_dec_architecture} we discuss linear dimension reduction methods (encoders and decoders) which are shared across all of the approaches that we consider here except for Fourier neural operator. In \Cref{sec:surr_neural} we consider neural operator architectures, namely reduced-basis neural operators and the Fourier neural operator, and consider two training formulations: one based solely on operator evaluations and another that additionally incorporates
(Fr\'echet) derivative information. In \Cref{sec:surr_poly} we present sparse polynomial surrogates and tensor-train surrogates. In \Cref{sec:approximation_theory} we review the relevant approximation-theoretic results for the surrogates that we consider here. Finally, in \Cref{sec:numerical_results} we report the results of the empirical comparisons.

\section{Encoders and decoders} \label{sec:enc_dec_architecture}

Let $\CX$, $\CY$ be two real separable Hilbert spaces and let $\CG$ be a mapping from (a subset of) $\CX$ to $\CY$. The input is assumed to be distributed according to a measure $\mu$ on $\CX$.

\subsection{Representation systems}

Since $\CX$ and $\CY$ are in general infinite-dimensional, in order to numerically approximate $\CG$ we must choose a representation system in which to express elements in these spaces. This can in principle be done by any finite linear or nonlinear parametrization. In this study, we focus on linear representations, which are a suitable choice whenever operator inputs and outputs can be well-represented in linear subspaces of moderate dimension, i.e., have small linear $n$-widths \cite{MR774404}. This is known to be valid in particular if the covariance of $\mu$ has fast decaying eigenvalues, and the operator $\CG$ is smooth \cite{CoDeSch1,CoDe}.

Fixing two orthonormal bases
\begin{equation}\label{eq:ONBs}
  (\psi_i)_{i\in\N}\subset\CX\qquad\text{and}\qquad (\eta_j)_{j\in\N}\subset\CY
\end{equation}
and shifts
\begin{equation*}
  m_\CX\in\CX,\qquad m_\CY\in\CY,
\end{equation*}
there exists a one-to-one relationship between elements in $x\in\CX$ and the coefficients of the shifted input, via the \emph{encoder}
\begin{subequations}\label{eq:ED}
\begin{align}
  \CE:\begin{cases}\CX\to\ell^2(\N)\\
         x\mapsto (\langle x-m_\CX,\psi_i\rangle_{\CX})_{i\in\N}
         \end{cases}
\qquad\text{and its inverse}
\qquad
  \CE^{-1}:\begin{cases} \ell^2(\N)\to \CX\\
             (c_i)_{i\in\N}\mapsto m_\CX+\sum_{i\in\N}c_i\psi_i
             \end{cases}.
\end{align}
Similarly, on the output side we introduce the \emph{decoder}
\begin{align}\label{eq:dec}
  \CD:\begin{cases}
  \ell^2(\N)\to \CY\\
        (c_j)_{j\in\N}\mapsto m_\CY+\sum_{j\in\N}c_j\eta_j
         \end{cases}
\qquad\text{and its inverse}\qquad
  \CD^{-1}:
  \begin{cases}
    \CY\to\ell^2(\N)\\
    y\mapsto (\langle y-m_\CY,\eta_j\rangle_{\CY})_{j\in\N}
  \end{cases}.
\end{align}
\end{subequations}
Additionally, we denote by $\CE_r$, $\CD_r$ the truncated versions of these operators, i.e., $\CE_r:\CX\to\R^r$ and $\CD_r:\R^r\to m_\CY+{\rm span}\{\eta_1,\dots,\eta_r\}$.
The truncated operators are not invertible, but we can define pseudoinverses
\begin{equation*}
    \CE_r^\dagger := m_{\CX} + \sum_{i=1}^r c_i \psi_i \qquad \text{and} \qquad \CD_r^\dagger := (\iprod{y - m_\CY, \eta_j})_{j=1}^r.
\end{equation*}

\subsection{Architecture and sparsity}

For any mapping $\CG:\CX\to\CY$, there exists a coefficient map $g:\ell^2(\N)\to\ell^2(\N)$ defined as
\begin{equation*}
  g:= \CD^{-1}\circ\CG\circ\CE^{-1},
\end{equation*}
such that
\begin{equation}\label{eq:DgE}
  \CG=\CD\circ g \circ\CE.
\end{equation}
Encoder-decoder architectures approximate $\CG$ via
\begin{equation}\label{eq:DgEtrunc}
  \CG\simeq \CD_{d_{\rm out}} \circ \tilde g\circ \CE_{d_{\rm in}},
\end{equation}
where $\tilde g:\R^{d_{\rm in}}\to\R^{d_{\rm out}}$.
For future reference we introduce
\begin{equation}\label{eq:gtruncref}
    g_{d_{\rm in}}^{d_{\rm out}} = \CD_{d_{\rm out}}^\dagger \circ \CG \circ \CE_{d_{\rm in}}^\dagger.
\end{equation}

\begin{figure}[ht]
  \centering
  \begin{tikzpicture}[scale=1, every node/.style={scale=1}]
      \node[draw, minimum size=2cm]   (A) at (0,0) {$K^s \subset \CX$};
      \node[draw, minimum size=1.5cm] (B) at (4,0) {$\substack{\displaystyle \bigtimes_{i \in \N} [-\lambda_i^s, \lambda_i^s] \\ \subset \ell^2(\N)}$};
      \node[draw, minimum size=1.5cm] (C) at (8,0) {$\ell^2(\mathbb{N})$};
      \node[draw, minimum size=2cm]   (D) at (12,0) {$\CY$};
      \draw[-latex, dashed]  (B.south east) to [bend right = 15] node[below] {$\tilde g$} (C.south west);
      \draw[-latex] (A.east) to[bend left] node[above] {$\CE$} (B.west);
      \draw[-latex] (B.west) to[bend left] node[below] {$\CEinv$} (A.east);
      \draw[-latex] (B.east) -- node[above] {$g$} (C.west);
      \draw[-latex] (C.east) to[bend left] node[above] {$\CD$} (D.west);
      \draw[-latex] (D.west) to[bend left] node[below] {$\CDinv$} (C.east);
      \draw[-latex] (A.north east) to[bend left=15] node[above] {$\CG$} (D.north west);
  \end{tikzpicture}
  \caption{Encoder-decoder architecture.}
  \label{diagram:encoder-decoder-expanded}
\end{figure}
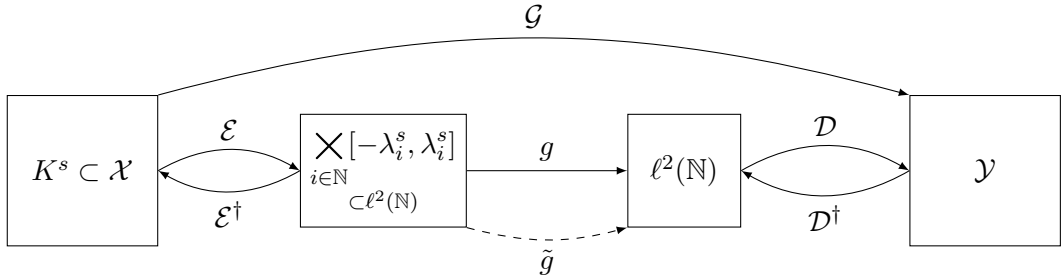

The truncation of the input after $d_{\rm in}$ coefficients, and the restriction of the output to an $d_{\rm out}$-dimensional subspace, induce an error in the overall approximation independent of $\tilde g$. As mentioned before, this approach can only be successful if there exist subspaces of $\CY$ and $\CX$ of suitable dimension, in which the input and output possess a good approximation. Mathematically, one way to control this error, is by assuming a certain decay in the importance of coefficients, quantified by a strictly positive, monotonically decreasing sequence $(\lambda_i)_{i\in\N}\in\ell^1(\N)$: specifically, following \cite{herrmann2024neural}, for $s\ge 0$ we define the Hilbert space $\CX^s\subseteq\CX$ via the inner product
\begin{equation}\label{eq:Xsinner}
  \langle x,\tilde x\rangle_{\CX^s}:=\sum_{i\in\N}\lambda_i^{-2s}\langle x,\psi_i\rangle_{\CX}\langle \tilde x,\psi_i\rangle_{\CX}.
\end{equation}
For Fourier (or wavelet) type bases, these spaces can be understood as smoothness spaces, such as Sobolev or Besov spaces, with $s$ controlling the smoothness parameter.
We will then consider the approximation of $\CG$ on subsets of the type
\begin{equation}\label{eq:Ks}
    K^s
    := m_\CX+\left\{
      \sum_{i \in \N} x_i \lambda_i^s \psi_i \middle\vert \forall i \in \N \colon x_i \in [-1,1] \right\},
\end{equation}
for some $s\ge 1/2$ (to ensure $\ell^2$ summability of the coefficient sequences). Note that if $m_\CX\in\CX^{s-1/2}$, then $K^s\subset \CX^{s-1/2}$.
The resulting encoder-decoder architecture is shown in \Cref{diagram:encoder-decoder-expanded}.

\subsection{Choice of basis and computation} \label{sec:ONB_choices}

In general, an optimal basis having minimal impact on the approximation error depends on the operator $\CG$, e.g.\ \cite{PC-ActSubspc}. Since this makes it hard to compute, a common choice is to rely on the best approximation of the underlying measure in terms of the least squares error. Recall that $x\sim\mu$ so that $y=\CG(x)\sim\CG_\sharp\mu$, with the latter denoting the pushforward of $\mu$ under the operator.
For a fixed $r\in\N$, and a set of orthonormal vectors $\{\psi_1,\dots,\psi_r\}$, denote by
\begin{equation}\label{eq:projection}
  \Pi_r^\psi:\begin{cases}
          \CX\to{\rm span}\{\psi_1,\dots,\psi_r\}\\
          x\mapsto \sum_{j=1}^r \langle x,\psi_j\rangle_{\CX}\psi_j,
        \end{cases}
\end{equation}
the orthogonal projection onto ${\rm span}(\psi_1,\dots,\psi_r)$. The goal is then to find $m_\CX$ and $\{\psi_1,\dots,\psi_r\}$ minimizing
\begin{equation}\label{eq:best_r_term_approximation}
    \mathbb{E}_{x \sim \mu} \left[ \bnorm{x-(m_\CX+\Pi^\psi_r(x-m_\CX))}_{\CX}^2 \right],
\end{equation}
i.e., the affine subspace, such that the projection onto this subspace minimizes the mean-square error.

Similarly, for the output space we wish to find $m_\CY$ and $\{\eta_1,\dots,\eta_{r}\}$ minimizing
\begin{equation*}
    \mathbb{E}_{y \sim \CG_\sharp\mu} \left[ \bnorm{y-(m_\CY+\Pi^\eta_{r}(y-m_\CY))}_{\CY}^2 \right]
    = \mathbb{E}_{x \sim \mu} \left[ \bnorm{\CG(x)-\Pi^\eta_r\CG(x)}_{\CY}^2 \right],
\end{equation*}
where $\Pi_r^\eta$ is defined analogous to \eqref{eq:projection}.

\subsubsection{PCA}

Since the procedure is the same for $\mu$ (to create the input representation system $(\psi_j)_j$) and $\CG_\sharp\mu$ (to create the output representation system $(\eta_j)_j$), we only focus on the former.

For any fixed $\{\psi_1,\dots,\psi_r\}$, the best constant approximation
of $x-\Pi_r^{\CX}x$ in $L^2_\mu$ is its expected value $\E_{x\sim\mu}[x-\Pi_r^\CX x]=\E_{x\sim\mu}[x]-\Pi_r^\CX \E_{x\sim\mu}[x]$. Therefore we may set $m_\CX=\E_{x\sim\mu}[x]$ in the following. By the Pythagorean Theorem
  \begin{equation*}
    \norm[\CX]{(x-m_\CX)-\Pi_r^\CX (x-m_\CX)}^2 = \norm[\CX]{x-m_\CX}^2 - \norm[\CX]{\Pi_r^\CX (x- m_\CX)}^2,
  \end{equation*}
  so that
\begin{align*}
    \mathbb{E}_{x \sim \mu} [\bnorm{x-(m_\CX+\Pi_r^\CX (x-m_\CX))}_{\CX}^2] =
    \mathbb{E}_{x \sim \mu} [\bnorm{x-m_\CX}_{\CX}^2- \bnorm{\Pi_r^\CX(x-m_\CX)}_{\CX}^2].
\end{align*}
Since the first term does not depend on $\{\psi_1,\dots,\psi_r\}$, an optimal solution is given as the one that maximizes the term
\begin{equation}\label{eq:maximizethis}
  \mathbb{E}_{x \sim \mu}\big[\|\Pi_r^\CX(x-m_\CX)\|_{\CX}^2\big].
\end{equation}

Let $C:\CX\to\CX$ be the covariance operator of $\mu$ defined via
\begin{equation*}
    \langle Cx,\tilde x\rangle_\CX
    = \int_{\CX} \langle x, z - m_\CX\rangle_\CX \langle \tilde x, z - m_\CX\rangle_\CX \,\dd\mu(z).
\end{equation*}
which is well-defined if $\mu$ has finite first and second moments.
It is well known that a maximizer of \eqref{eq:maximizethis} is attained by an $r$-dimensional subspace spanned by the eigenvectors of $C$ corresponding to $r$ of its largest eigenvalues.
This can be seen by noting that
\begin{align*}
    \mathbb{E}_{x \sim \mu} [\bnorm{\Pi_r^\CX (x-m_\CX)}_{\CX}^2]
    &= \int_{\CX} \bnorm{\Pi_r^\CX (x-m_\CX)}_{\CX}^2 \dd \mu(x)\\
    &= \sum_{i=1}^r \int \iprod{\Pi_r^\CX (x-m_\CX), \psi_i}_{\CX}^2 \dd \mu(x)
    = \sum_{i=1}^r \iprod{\psi_i, C \psi_i}_{\CX}.
\end{align*}
For more details, see e.g., \cite{ramsay1997functional}.

\begin{remark}
This construction is closely connected to Karhunen--Lo\`eve expansions, which are commonly used in uncertainty quantification and operator learning to represent random fields. A Karhunen--Lo\`eve expansion expands a (mean-zero) random variable with distribution $\mu$ in the eigenfunctions of its covariance operator, which gives an (optimal) $L^2_\mu$ representation.
\end{remark}

\subsubsection{Empirical PCA}

In practice, we consider the discrete, empirical case, i.e., elements of $\CX$ are discretized with a finite number $m \in \N$ of (real-valued) degrees of freedom
and we only have a finite number $N \in \N$ of samples.
Let $X=[x_1,\dots,x_N] \in \R^{m \times N}$ contain these samples as columns.
For our application, we want orthogonality with regard to the inner product on the Hilbert space $\CX$ (or $\CY$ in case of the output).
Assuming that $\CX$ is discretized using a linear basis $(\chi_i)_{i=1, \dots, m} \subset \CX$ such as finite elements,
we can compute $M = (\iprod{\chi_i, \chi_j}_{\CX})_{i, j = 1}^{m} $.
This matrix represents the discrete representation of the Riesz map w.r.t.\ our basis (the Gram matrix). It is thus symmetric positive definite and admits a Cholesky decomposition $M = R^T R$.
Additionally let $\bar x=\frac{1}{N}\sum_{j=1}^N x_j$ be the sample mean and
$\tilde X = [\tilde x_1,\dots,\tilde x_N]\in\R^{m\times N}$ with $\tilde x_i= x_i-\bar x$ the centered samples.
We then have
\begin{equation*}
    \sum_{i=1}^N \tilde x_i^T M \tilde x_i
    = \operatorname{tr}(\tilde X^T M \tilde X)
    = \operatorname{tr}(\tilde X^T R^T R \tilde X)
    = \bnorm{R\tilde X}_{{\rm F}}^2,
\end{equation*}
where $\|\cdot\|_{\rm F}$ denotes the Frobenius norm.

It is a classical result \cite{EckartYoung1936, Mirsky1960} that best rank-$r$ approximation to a matrix with regard to the Frobenius norm (or any unitarily invariant norm) is provided by truncated SVD.
More precisely, given $R\tilde X = U \Sigma V^T$ with unitary $U,V$ and diagonal $\Sigma$, we obtain a best rank-$r$ approximation as $U \Sigma_r V^T$, where $\Sigma_r$ is obtained from $\Sigma$ by setting all but the first $r$ diagonal entries to $0$.
The corresponding coefficients of the optimal basis are then given by the first $r$ columns of $R^{-1} U$, which is orthonormal with respect to the discrete inner product $M$ by construction.
Equivalently, we can also compute the eigen decomposition $\tilde X^T M \tilde X = V \Sigma^2 V^T$ and obtain $R^{-1} U = \tilde X V \Sigma^{-1}$.
Note that $\tilde X^T M \tilde X$ is the matrix representation of the
covariance operator $C$ in the computational basis induced by
$(\chi_i)_{i=1}^m$, so that its eigenpairs correspond to the principal
components.

\section{Neural operator surrogates} \label{sec:surr_neural}

In our experiments, we focus on two types of neural operator surrogates: Reduced-basis neural operators (RBNOs) and Fourier neural operators (FNOs). We shortly recall their architecture, and afterwards describe training processes.

\subsection{Reduced-basis neural operators}\label{sec:pcanet}

Suppose data given in the form of operator input-output pairs $(x_i,\CG(x_i))_{i=1}^n\subset \CX\times\CY$. The fundamental idea of reduced-basis neural operators (RBNOs) is to encode operator inputs and outputs into low-dimensional coefficient representations that are independent of the underlying mesh or discretization, and to learn the resulting coefficient-to-coefficient map using a (typically fully connected) feedforward neural network. In terms of \eqref{eq:DgE}, this means that $\CG_{\rm RB-NO} = \CD \circ \tilde g \circ \CE$, where $\CE:\CX\to\R^{d_{\rm in}}$ and $\CD:\R^{d_{\rm out}}\to\CY$ are computed a priori, e.g. using empirical PCA on the data $(x_i)_{i=1}^n$, $\CG(x_i)_{i=1}^n$ respectively. The neural network $\tilde g:\R^{d_{\rm in}}\to\R^{d_{\rm out}}$ is then trained to approximate the coefficient-to-coefficient map $g$. When choosing a fully connected feedforward architecture, the network is a composition of affine maps and element-wise nonlinearities. More precisely,
\begin{equation*}
  \tilde g= A_L\circ\sigma\cdots \sigma\circ A_1\circ\sigma\circ A_0,
\end{equation*}
where $\sigma:\R\to\R$ is an activation function, applied element-wise. The $A_\ell:\R^{d_{\ell}}\to\R^{d_{\ell+1}}$ are affine linear maps, i.e., $A_\ell (x) = W_\ell x + b_\ell$ where $W_{\ell} \in \R^{d_{\ell+1} \times d_{\ell}}$ is a weight matrix and $b_{\ell} \in \R^{d_{\ell+1}}$ is a bias vector. The entries of the weight matrices and bias vectors make up the trainable parameters of the architecture. We have $d_0=d_{\rm in}$ and $d_{L+1}=d_{\rm out}$, while the depth $L$ and layer widths $d_j$ can be adjusted to tune the desired expressive power of the network and its inductive bias.

RBNOs using orthonormal bases to represent inputs and outputs have been proposed and analyzed by various authors and under several names, e.g., \citep{hesthaven2018non} (\textit{POD-NN}), \citep{bhattacharya2021model, lanthaler2023operator} (\textit{PCA-Net}),
\citep{o2022derivative,o2022learning} (derivative-informed reduced basis architectures), \cite{Fanaskov2023} (\emph{Spectral neural operators}), \citep{herrmann2024neural, reinhardt2024statistical} (in a generalized formulation as \textit{FrameNet}). The primary advantage of these architectures is that training occurs in a reduced coefficient space, so the computational complexity does not scale with the ambient discretization dimension. In addition, they provide a simple and interpretable representation of the dominant input and output modes, which can be readily recomputed across different discretizations of the same problem. However, linear dimension reduction strategies do not lead to as expressive architectures as those utilizing fully nonlinear representations of the input and output functions.

\subsection{Fourier neural operator}

The Fourier neural operator (FNO) \citep{li2021fourier} learns input–output maps through compositions of global spectral convolution operators in Fourier space together with pointwise nonlinearities. While the spectral convolution layers operate on truncated Fourier representations, the architecture also includes pointwise linear transformations of the feature field in physical space, and is therefore not strictly band-limited in the same sense as PCA-Net. This structure enables efficient learning of Green’s-function–like operators via the truncated Fourier representation while still allowing high-frequency information to be represented through the local channels and nonlinearities. The two key operations of the architecture are resolution-independent on uniform grids, allowing the same model to be applied across different refinements of the same discretization. Various adjustments, such as more sophisticated architectures, low rank tensor compression of the weights and adaptations to non-uniform grids, have since been proposed, e.g., \cite{kossaifi2023multigridtensorizedfourierneural,li2023fourier}.

\subsubsection{Continuous view}

FNO was originally introduced in its ``continuous form'', i.e., without spatial discretization. While computationally not feasible, this viewpoint turns out to be valuable, as it helps to reason about the behavior of the architecture in the limit of infinite resolution.
To explain the construction consider the $d$-dimensional unit square $\Omega= [0,1]^d$, and assume that $\CX$ and $\CY$ are function spaces over $\Omega$. The operator $\CG$ thus takes a function $x:\Omega\to\R^{d_x}$ to another function $y:\Omega\to\R^{d_y}$.

The FNO architecture includes a range of hyperparameters, in particular the channel space width $\nu$, depth $L$, maximum number of Fourier modes $k$, and nonlinear activation function $\sigma$.
For a given choice of the parameters, the FNO maps $x\in\CX$ to
\begin{equation}\label{eq:FNOarch}
\CG_{\rm FNO} (x) := Q \circ \sigma \circ M_L \circ \sigma \circ M_{L-1}  \circ  \dots \circ M_1 \circ P (x)\in\CY.
\end{equation}
Here $P : C^0(\Omega,\R^{d_x}) \to C^0(\Omega,\R^\nu)$ is a lifting layer, $Q : C^0(\Omega,\R^\nu) \to C^0(\Omega,\R^{d_y})$ is a projection layer, and $M_\ell:C^0(\Omega,\R^{\nu})\to C^0(\Omega,\R^{\nu})$ are the Fourier layers.
The lifting and projection layers are implemented as linear transformations represented by matrices in $\R^{\nu\times d_x}$ and $\R^{d_y\times \nu}$, which are applied pointwise in the spatial coordinates to obtain the desired number of channels.
The application of the nonlinear activation function $\sigma:\R\to\R$ is understood componentwise. We emphasize that \eqref{eq:FNOarch} corresponds to composing the input function $x$ with several other functions.

Let us now describe the Fourier layers. Each $M_\ell$ takes in a function $v\in C^0(\Omega,\R^\nu)$ and returns the function
\begin{equation*}
  \mathcal{F}_k^{-1} \left( V_\ell \mathcal{F}_k(v) \right)(\cdot) + W_\ell v(\cdot)
\end{equation*}
in $C^0(\Omega,\R^\nu)$. Here $\mathcal{F}_k : C^0(\Omega, \R^\nu) \to \R^{k \times \nu}$ and $\mathcal{F}_k^{-1} : \R^{ k \times \nu} \to C^0(\Omega, \R^\nu)$ are the truncated Fourier transform and its (zero padded) inverse. The map $V_\ell: \R^{k \times \nu} \to \R^{k \times \nu}$ is a linear transformation in the (truncated) frequency domain represented by a tensor $\R^{k \times \nu \times \nu}$. Specifically, for each $m\in\{1,\dots,k\}$ (representing the $m$th Fourier mode), the linear transformation $V_{\ell,m}\in\R^{\nu\times\nu}$ is applied to the $m$th row of the input. This ensures that there is no mixing between Fourier modes, but there is exchange of information within the channels of each Fourier mode. Finally, $W_\ell:C^0(\Omega,\R^\nu) \to C^0(\Omega,\R^\nu)$ is a linear transformation represented by a matrix in $\R^{\nu \times \nu}$, which is again applied pointwise in the spatial domain $\Omega$.

\subsubsection{Discrete view and implementation}\label{ssec:fno_discrete}

In practice, Fourier transformations cannot be computed exactly. Instead, they are replaced by performing a Fast Fourier Transformation (FFT) on a $d$-dimensional grid.

Although non-standard, the architecture can again be interpreted as a special case of \eqref{eq:DgEtrunc}, where the encoder evaluates $x\in C^0(\Omega,\R^{d_x})$ on uniform grid points $0\le t_1<\dots<t_n\le 1$ and then applies the linear transformation $P$, i.e.,
\begin{equation}
  \CE:\begin{cases}
        C^0(\Omega,\R^{d_x})\to \R^{n\times\cdots\times n\times \nu}\\
        x\mapsto P((x(t_{i_1},\dots,t_{i_d})_j)_{i_1,\dots,i_d=1}^n)_{j=1}^{d_x}.
      \end{cases}
\end{equation}
The application of the linear transformations $V_\ell$, $W_\ell$ remains the same, but acting on vectors. The Fourier transform $\CF_k$ is replaced by its (fast) discrete counterpart. This, in general, necessitates rectangular domains $\Omega$ and uniform grids. Finally, the decoder can be interpreted as the map
\begin{equation*}
Q\circ\sigma\circ M_L,
\end{equation*}
where the inverse Fourier transform within $M_L$ essentially does an expansion in the Fourier basis, which is followed by the activation function and a final linear transformation. We emphasize that this viewpoint (especially in terms of what belongs to encoding, decoding, and coefficient mapping) for FNO is not canonical and up to interpretation.

\subsection{Training in $L_\mu^2$ and $H_\mu^1$}\label{sec:training_L2_H1}

Suppose again given samples $(x_i,\CG(x_i))_{i=1}^N$, where $x_i\sim\mu$ are assumed i.i.d. draws. A standard way of training the above type of architecture is to minimize the mean square error
\begin{equation}\label{eq:L2obj}
\frac{1}{N} \sum_{j=1}^N\|\CG(x_j)-\tilde\CG(x_j)\|^2_{\CY}\simeq \E_{x \sim \mu} \left[\|\CG(x)-\tilde\CG(x)\|_\CY^2 \right]=\|\CG-\tilde\CG\|_{L^2_\mu}^2,
\end{equation}
over the operators $\tilde\CG$ within our surrogate class. For architectures as in \eqref{eq:DgEtrunc}, the encoders and decoders can be trained simultaneously, or determined in a first step (as e.g., when using PCA).

Training in $L^2_\mu$ may in practice be insufficient when the operator surrogate is used in downstream tasks that require accurate derivatives. Typical examples arise in optimal control or optimal experimental design, where the objective is to satisfy stationary conditions for an objective functional. These conditions take the form of variational inequalities involving the gradient of the objective functional. To improve performance in these situations, the $L^2_\mu$ objective in \eqref{eq:L2obj} can be adjusted to include derivatives, and instead train in Sobolev type spaces such as $W^{k,p}_\mu$, \cite{OLEARYROSEBERRY2024}. Specifically, for the common case $k=1$, $p=2$, the objective becomes
\begin{equation}\label{eq:H1obj}
\E_{x\sim\mu}[\|\CG(x)-\tilde\CG(x)\|_\CY^2] +\E_{x\sim\mu}[\|D\CG(x)-D\tilde\CG(x)\|^2]
\end{equation}
where $D\CG$, $D\tilde\CG\in L(\CX,\CY)$ denote the Fr\'echet derivatives (assuming they exist). The training formulation depends on a suitable norm for the derivative error, e.g., the operator norm or Hilbert--Schmidt norm.
This formally corresponds to the training in the $H^1_\mu$ parametric Bochner (Sobolev) space, and in the rest of this section, we discuss practical and theoretical aspects in a bit more detail. First, we point out, however, that the approximation of \eqref{eq:H1obj} requires samples also of the derivatives, e.g., $(x_i,\CG(x_i),D\CG(x_i))_{i=1}^N$; therefore, generating data for derivative-informed training is more expensive. However, as discussed in various works \cite{OLEARYROSEBERRY2024,cao2025derivative}, it can be made marginally inexpensive relative to forward PDE solves by amortizing computations (e.g., factorizations) associated with the forward simulation. $H^1_\mu$ operator learning has been used extensively in reduced basis architectures \cite{luo2025efficient,cao2026lazydino} as well as FNO \cite{yao2025derivative}.

\subsubsection{Generation of derivative samples for PDEs}\label{sec:derivative_sample_gerneation}

Let $\CR:\CX\times\CY\to\CY'$ denote a nonlinear PDE operator, such that the operator $\CG:\CX\to\CY$ is implicitly defined as $\CR(x,\CG(x))=0$.
Then, assuming the solution map is Fr\'echet differentiable with respect to $x$,
\begin{equation*}
\frac{\partial}{\partial x}\CR(x,\CG(x)) +
\frac{\partial}{\partial y}\CR(x,\CG(x)) D\CG(x)=0,
\end{equation*}
so that
\begin{equation}\label{eq:DCGx}
D\CG(x) = -\underbrace{\Big(\frac{\partial}{\partial y}\CR(x,\CG(x))\Big)^{-1}}_{\in L(\CY',\CY)}\;\underbrace{\frac{\partial}{\partial x}\CR(x,\CG(x))}_{\in L(\CX,\CY')}\in L(\CX,\CY).
\end{equation}
The term $\frac{\partial}{\partial y}\CR$ is the linearized PDE operator. Thus, to compute $D\CG(x)$, in practice, another PDE needs to be solved. In case the original PDE was linear, they are the same. The next example shows this.

\begin{example}
  Let $\CX=H^1([0,1])$, let $\CY=H_0^1([0,1])$, and consider the map $\CG:x\mapsto y$ implicitly defined via $\CR(x,y)=0$ where
  \begin{align*}
    \CR(x,y) &= \nabla \cdot (x\nabla y)+f &&\text{in }[0,1]\\
               y&=0 &&\text{on }\{0,1\},
  \end{align*}
  for some $f\in \CY'=H^{-1}([0,1])$. Then
  \begin{equation*}
    \frac{\partial}{\partial x}\CR(x,y)(h)=\nabla\cdot(h\nabla y)
    \quad\text{and}\quad
    \frac{\partial}{\partial y}\CR(x,y)(k)=\nabla\cdot(x\nabla k)
    \qquad\forall h\in\CX,~k\in \CY.
  \end{equation*}
  In particular, by \eqref{eq:DCGx}, the Fr\'echet derivative $D\CG(x)h$ of the operator $\CG$ at $x\in\CX$ in direction $h\in\CX$ is the solution $z\in\CY$ of the PDE
  \begin{align*}
    -\nabla \cdot (x\nabla z)&=\nabla \cdot (h\nabla y) &&\text{in }[0,1]\\
    z&=0 &&\text{on }\{0,1\},
  \end{align*}
  where we assume that $x$ is uniformly positive such that the PDE is well-posed.
\end{example}
Additionally, when the forward PDE is nonlinear and solved using a Newton iteration, the same linearized PDE operator is used at each Newton step. In this case, the cost of computing derivatives can be even lower than in the linear forward problem.

\subsubsection{Reduction to coefficient derivatives}\label{sec:dertraining}

In practice, dealing with the Fr\'echet derivative of $\CG$, which is a linear mapping from $\CX\to\CY$, requires discretization. In the encoder/decoder setting, it is natural to use the coefficient representation $g$ in \eqref{eq:DgE}. In terms of the coefficients w.r.t.\ the ONBs $(\psi_j)_{j\in\N}$, $(\eta_j)_{j\in\N}$, the Fr\'echet derivative can be represented by the infinite matrix
\begin{equation*}
  \begin{pmatrix}
    \langle D\CG(x)\psi_1,\eta_1\rangle_{\CY} &\langle D\CG(x)\psi_2,\eta_1\rangle_{\CY} &\dots\\
    \langle D\CG(x)\psi_1,\eta_2\rangle_{\CY} &\ddots&\\
                                              \vdots &&
  \end{pmatrix}.
\end{equation*}

The weakest, canonical norm to measure the Fr\'echet derivative in is the operator norm of $D\CG(x)$. This corresponds to the spectral norm of the above infinite matrix. More practical and cheaper to compute is the (stronger) Hilbert-Schmidt norm. It corresponds to the Frobenius norm of the matrix representation, i.e.,
\begin{equation*}
  \|D\CG(x)\|_{{\rm HS}(\CX,\CY)}^2=\sum_{i,j\in\N}\langle D\CG(x)\psi_i,\eta_j\rangle_{\CY}^2.
\end{equation*}
In general, the bounded linear map $D\CG(x)$ need not belong to ${\rm HS}(\CX,\CY)$ however. Since we approximate the map on sets of type $K^s\subseteq \CX^{s-1/2}$ as introduced in \eqref{eq:Ks}, it is natural to also restrict the derivative approximation to some subspace $\CX^{\tilde s}$ for some $\tilde s\ge 0$ such as $\tilde s=s-1/2$. We then approximate the derivative in the (weaker) norm of ${\rm HS}(\CX^{\tilde s},\CY)$. By definition of the inner product \eqref{eq:Xsinner} in $\CX^{\tilde s}$, an orthonormal basis of $\CX^{\tilde s}$ is given by
\begin{equation*}
  \lambda_i^{\tilde s}\psi_i,~i\in\N.
\end{equation*}
Hence
\begin{equation*}
    \|D\CG(x)\|_{{\rm HS}(\CX^{\tilde s},\CY)}^2=\sum_{i,j\in\N}\langle D\CG(x)\psi_i,\eta_j\rangle_{\CY}^2\lambda_i^{2\tilde s}.
\end{equation*}

Let us write this in terms of the coefficient functions $g:\ell^2(\N) \to \ell^2(\N)$ in \eqref{eq:DgE}, i.e., where $\CG=\CD\circ g\circ\CE$ and with $\CE$, $\CD$ in \eqref{eq:ED}.
Since $\psi_i$, $\eta_j$ are ONBs,
\begin{equation*}
\langle D\CG(x)\psi_i,\eta_j\rangle_{\CY} = \partial_{i} g_j(\CE(x)).
\end{equation*}
Therefore
\begin{equation*}
    \|D\CG(x)\|_{{\rm HS}(\CX^{\tilde s},\CY)}^2=\sum_{i,j\in\N}(\lambda_i^{\tilde s}\partial_{i}g_j(\CE(x)))^2.
\end{equation*}

Finally, the surrogate $\tilde\CG=\CD_{d_{\rm out}}\circ\tilde g\circ \CE_{d_{\rm in}}$ in \eqref{eq:DgEtrunc}, acts on the truncated coefficient sequences. Up to a trunctation error, for encoder-decoder architectures \eqref{eq:DgEtrunc}, the second term in \eqref{eq:H1obj} w.r.t.\ $\|\cdot\|_{{\rm HS}(\CX^{\tilde s},\CY)}$ can thus be approximated via
\begin{equation}\label{eq:H1objtildeg}
\frac{1}{N}\sum_{k=1}^N \sum_{i=1}^{d_{\rm in}}\sum_{j=1}^{d_{\rm out}}\lambda_{i}^{2 \tilde s}\big(\partial_i g_j(\CE(x_k)) - \partial_i \tilde g_j(\CE_{d_{\rm in}}(x_k))\big)^2.
\end{equation}
The truncation error depends on $g$ and the truncation dimensions $d_{\rm in}$, $d_{\rm out}$, but is independent of $\tilde g$, over which the objective is minimized. Thus, the truncation error is not directly relevant for the training process, which motivates the minimization of \eqref{eq:H1objtildeg} in $\tilde g$. We refer to \cite{luo2025dimension} for further details.

\begin{remark}
Under suitable smoothness assumptions on $\CG$ and for $\tilde s$ sufficiently large, the norm $\|D\CG(x)\|_{{\rm HS}(\CX^{\tilde s},\CY)}$ can typically be shown to be finite, e.g., \cite{herrmann2024neural}.
\end{remark}

\section{Polynomial operator surrogates} \label{sec:surr_poly}

Consider the encoder-decoder architecture \eqref{eq:DgEtrunc}. In \Cref{sec:pcanet} we used a neural network to learn a suitable coefficient map $\tilde g:\R^{d_{\rm in}}\to\R^{d_{\rm out}}$ (typically, $\tilde g$ is only required on a subset of $\R^{d_{\rm in}}$). In this section, we will use multivariate polynomials to represent $\tilde g$ in \eqref{eq:DgEtrunc}.

Polynomial approximation and interpolation are classic and well-understood techniques, but naive generalizations to high dimensions suffer from the curse of dimensionality.
In this section, we first briefly recall polynomial interpolation, and subsequently introduce sparse-grid (SG) interpolation and tensor-trains (TT), and show how they can be used as tools for operator surrogates.

\subsection{Polynomial interpolation}\label{sec:polint}

\paragraph{Univariate interpolation.}

Let $\Omega \subset \R$ be an interval and for $\ell \in \N$ let $(\xi^\ell_i)_{i=0}^\ell \subset \Omega $ be $\ell + 1$ pairwise distinct interpolation points. Denote\footnote{We use in this section $c$ to denote the variable of the function to avoid a notational clash with the operator input $x\in\CX$, and to emphasize that $c$ is interpreted as an encoded \emph{coefficient}.} $\bbP_\ell := {\rm span} \set{c\mapsto c^i}{i=0,\dots,\ell}$. The polynomial interpolation operator $I^\ell : C^0(\Omega, \R) \to  \bbP_\ell$ maps a function $g\in C^0(\Omega,\R)$ onto the (unique) polynomial $I^\ell [g]$ of degree $\ell$ satisfying $g(\xi^\ell_i) = I^\ell [g](\xi^\ell_i)$ for all $i\in\{0,1,\dots,\ell\}$. Crucial for obtaining good interpolation error bounds is the choice of interpolation points with a small Lebesgue constant ${\rm Leb}((\xi^\ell_i)_{i=0}^\ell)$, which is defined as the operator norm of $I^\ell$,
$${\rm Leb}((\xi^\ell_i)_{i=0}^\ell) := \sup_{\norm[\infty]{p} = 1} \norm[\infty]{I^\ell[p]},$$
where $\norm[\infty]{\cdot}$ stands for the supremum norm. The approximation error can be bounded by
\begin{equation*}
    \bnorm{g - I^\ell[g]}_{\infty} \leq \left( 1+{\rm Leb}((\xi^\ell_i)_{i=0}^\ell) \right) \min_{p \in \bbP_\ell} \bnorm{g - p}_\infty,
\end{equation*}
i.e., the Lebesgue constant measures how close the interpolant is to the best approximating polynomial.
For interpolation on $\Omega =[-1,1]$, it can be shown that optimal points have a Lebesgue constant that grows as $\cO(\log \ell)$, with, e.g., Chebyshev points achieving this rate \cite{zbMATH01083107}.
If a nested sequence is desired, Leja sequences can be shown to have a Lebesgue constant that grows at most as $\cO(\ell^2 \log \ell)$ \cite{CHKIFA2013176, CALVI2011608}.
For interpolation on $\Omega = \R$, points satisfying suitable stability results exist as well, e.g.,\ Gauss-Hermite points \cite{doi:10.1137/17M1123079}.

One way to compute the interpolating polynomial known for its efficiency and numerical stability is barycentric interpolation, e.g., \citep{berrut2004barycentric}. The univariate barycentric interpolation formula is given as
\begin{align*}
    I^\ell [g] (c) := \sum_{i=0}^\ell \CL_i^\ell(c) g(\xi^\ell_i),
    \qquad \text{ with }
    \CL_i^\ell(c) &:= \left. \left( \frac{\omega_i^\ell}{c - \xi_i^\ell} \right) \middle/ \left(\sum_{k=0}^{\ell} \frac{\omega_k^\ell}{c - \xi_k^\ell} \right) \right. \\
    \quad \text{ and }
    \omega_i^\ell &:= \prod \limits_{k\in\{0,1,...,\ell\}/\{i\}}\frac{1}{(\xi^\ell_i - \xi^\ell_k)}.
\end{align*}
After pre-computing the barycentric weights $(\omega_i^\ell)_{i=0}^\ell$ in $\cO(\ell^2)$ operations, the interpolant can be evaluated using only $\cO(\ell)$ operations.

\paragraph{Tensorized interpolation.}

Let $d\in \N$. Given a tensor-product domain $\Omega = \otimes_{j=1}^d \Omega_j$ with $\Omega_j \subset \R$ and a multi-index $\bsnu \in \N_0^d$ characterizing the maximal polynomial degree in each dimension, we define $I^\bsnu : C^0(\Omega, \R) \to \bbP_\bsnu := {\rm span} \set{\bsc\mapsto\bsc^\bsmu}{\bsmu \leq \bsnu}$ as
\begin{equation*}
    I^\bsnu := \otimes_{j=1}^d I^{\nu_j}.
\end{equation*}
The tensorized barycentric interpolation formula can be expressed as
\begin{equation} \label{eq:ip_tensorproduct}
    I^\bsnu [g] (\bsc)
    = \sum_{j_1 = 0}^{\nu_1} \dots \sum_{j_d =0}^{\nu_d} \prod_{k=1}^{d} \CL_{j_k}^k (c_k) \Bv(j_1, \dots, j_d),
  \qquad \bsc \in \Omega
\end{equation}
with a tensor $\Bv \in \R^{\bsnu+{\bm 1}} := \R^{(\nu_1+1) \times (\nu_2+1) \times \dots \times (\nu_d+1)}$ given as
\begin{equation}\label{eq:collocation_tensor}
    \Bv(j_1, \dots, j_d) = f(\xi_{j_1}^1, \dots, \xi_{j_d}^d)
\end{equation}

Storing the tensor $\Bv$ and evaluating the interpolant \eqref{eq:ip_tensorproduct} both scale with $\cO\left(\textstyle \prod_{j=1}^d (1+\nu_j)\right)$, which is in general exponential in $d$ and thus quickly becomes prohibitive for moderate to high $d$ (curse of dimensionality). For this reason, direct tensorized interpolation is in general not computationally tractable in high dimensions.

\subsection{Reduced-basis sparse-grid surrogates}

The core idea of sparse-grid polynomial interpolation is to reduce the complexity by considering a combination of tensorized interpolation operators whose total number of degrees of freedom remains bounded. This methodology can be applied to approximate the coefficient map $g$ in \eqref{eq:DgE}, leading to a reduced-basis sparse-grid surrogate as proposed in \cite{herrmann2024neural}.

\subsubsection{Sparse-grid interpolation}

Let $d_{\rm in}\in\N$, and for $\bsc\in\R^{d_{\rm in}}$, consider the  polynomial ansatz spaces $\bbP_\Lambda := {\rm span} \set{\bsc\mapsto\bsc^\bsmu}{\bsmu \in \Lambda}\subseteq C^0(\R^{d_{\rm in}})$ parametrized by downward closed multi-index sets $\Lambda \subset \N_0^{d_{\rm in}}$. The interpolation operator
\begin{equation} \label{eq:ip_smolyak}
    I^\Lambda := \sum \limits_{\bsnu \in \Lambda} \zeta_{\Lambda, \bsnu} I^\bsnu, \qquad \zeta_{\Lambda, \bsnu} := \sum \limits_{\bse \in \{0,1\}^d : \bsnu+\bse \in \Lambda} (-1)^{|\bse|}
\end{equation}
is exact on $\bbP_\Lambda$, e.g., \cite[Theorem 2.1]{Chkifa2014}.
Under certain algebraic sparsity assumptions, \cite{CoDeSch1,CoDe}, typical ansatz spaces tailored to high-dimensional but smooth target functions are parametrized by multi-index sets of the form
\begin{equation}\label{eq:ip_lambda}
    \Lambda_{a, b, \ell} := \Big\{\bsnu \in \N_0^d \ : \  \sum_{j=1}^d \log(a + j b) \nu_j < \ell\Big\},
\end{equation}
see for instance \cite{BCM17,JZdiss}. The parameter $\ell > 0$ controls the cardinality of the ansatz space, while $a > 1$ and $b > 0$ control the distribution of degrees of freedom over the dimensions.
Here we only mention that if the target function $g$ stems from a coefficient representation of a smooth operator as in \eqref{eq:DgE}, then the logarithmic increase of weights can be justified in case the encoded input coefficients $\CE(x)$ decrease algebraically; this can be made rigorous for instance if $(\psi_i)_{i\in\N}$ in \eqref{eq:ONBs} is a Fourier basis, and $x$ belongs to a Sobolev or Besov type function space \cite{herrmann2024neural}.

\subsubsection{Vector-valued sparse-grid interpolation}\label{sec:vectorSmolyak}

Since we wish to represent the function $\tilde g:\R^{d_{\rm in}}\to\R^{d_{\rm out}}$, we need to interpolate a vector-valued function. This is straightforward, by interpolating each of its $d_{\rm out}$ components. The analysis in \citep{herrmann2024neural} employs different interpolation operators $I^{\Lambda_j}$ for each output dimension $j\in\{1,\dots,d_{\rm out}\}$, where
\begin{equation*}
  \Lambda_1\supseteq\Lambda_2\supseteq\dots\Lambda_{d_{\rm out}}.
\end{equation*}
In theory, this allows for more efficient evaluations. In practice, we choose to use $\Lambda_1=\dots=\Lambda_{d_{\rm out}}$ equal across all output dimensions for two reasons:
\begin{itemize}
  \item[(i)]
  Decreasing ansatz spaces does not have a positive impact on the \textit{offline} cost of constructing the operator surrogates. When using a nested set of interpolation points, the overall number of operator evaluations required is only determined by the largest ansatz space $\Lambda_1$. For non-nested interpolation points using smaller ansatz spaces in higher dimensions would even require \textit{additional} operator evaluations.
  \item[(ii)] Smaller ansatz spaces in higher dimensions can, in principle, reduce the \textit{online} computational cost of evaluating the operator surrogate by allocating less effort to dimensions with minimal contributions to the overall surrogate. However, modern high-performance software and hardware are optimized for processing large volumes of identically structured data, which may offset the benefits of such tailored ansatz spaces.
\end{itemize}

\subsubsection{Operator surrogate and training}

To obtain a surrogate of an operator $\CG$, we use the sparse-grid interpolant to interpolate the components of $g:\R^\N\to\R^\N$ in \eqref{eq:DgE}. Specifically, after fixing $d_{\rm in}$, $d_{\rm out}\in\N$ and a finite downward closed index set $\Lambda\subseteq\N_0^{d_{\rm in}}$, for example as in \eqref{eq:ip_lambda}, we define with $\bsc=(c_1,\dots,c_{d_{\rm in}})\in\R^{d_{\rm in}}$
\begin{equation*}
  \tilde g_j := I^{\Lambda}[\bsc\mapsto g_j(\bsc,0,0,\dots)]\in\bbP^\Lambda\qquad\forall j=1,\dots,d_{\rm out}.
\end{equation*}
The operator surrogate is then given by \eqref{eq:DgEtrunc}, i.e, $\CG_{\rm RB-PI} = \CD \circ \tilde g \circ \CE$.

A major advantage of this approach is that no nonconvex optimization is required. This makes the method robust and easily reproducible. A disadvantage is that the values
\begin{equation*}
  g_j(\bsc,0,0,\dots) = \Big\langle \CG\big(\sum_{i=1}^{d_{\rm in}}c_i\psi_i\big),\eta_j\Big\rangle_{\CY}
\end{equation*}
are required for all interpolation $\bsc$ needed for the computation of $I^\Lambda$. Thus, the method is not feasible if $\CG$ is only known at random inputs, but requires (at least approximate) probing of $\CG$ at arbitrary points. If this is not the case, the interpolation can, however, be replaced by least squares, see \cite{cohen2013on, cohen2017optimal, chkifa2015discrete}. Note, however, that the accuracy and data efficiency of least squares approaches also benefit when evaluating the target operator at points sampled from a judiciously chosen distribution. For an adaptation of the weighted least squares approach to the context of operator learning, see \cite{turnage2025optimal}.

\subsection{Reduced-basis tensor-train surrogates}\label{sec:TT}

Instead of directly constraining the polynomial space, another possibility is to reduce the computational and storage complexity by using an efficient approximation to the tensor $\Bv$ of collocation points in \eqref{eq:collocation_tensor}.

\subsubsection{Tensor-trains}

Consider $\bsv \in \R^{\bsnu +1}$ as found in \eqref{eq:collocation_tensor}.
While the full tensor has storage complexity $\cO(\textstyle \prod_{k=1}^d (1+\nu_k))$,
compressed tensor formats exist to store it more efficiently, assuming that the tensor is of low rank in a suitable way. A popular format is the \emph{tensor-train (TT)} format \citep{osedelets11}, where tensors take the form
\begin{equation*}
    \Bv(j_1 \dots j_d)
    = \sum_{\alpha_1, \dots, \alpha_{d-1}=1}^{r_1, \dots, r_{d-1}}
        \Bv^{(1)}_{\alpha_1}(j_1)
        \cdots \Bv^{(k)}_{\alpha_{k-1}, \alpha_{k}}(j_k)
        \cdots \Bv^{(d)}_{\alpha_{d-1}}(j_d)
    \qquad \forall k \colon 0 \leq  j_k \leq \nu_k.
\end{equation*}
The 3-tensors $\Bv^{(k)} \in \R^{r_{k-1} \times (\nu_k + 1) \times r_k}$ are the \emph{(TT) cores} and
$r_0, \dots, r_d \in \N$ the \emph{TT ranks} of $\Bv$, where by convention $r_0 = r_d = 1$.
Such a TT tensor requires storing $\cO(\textstyle d (\nu_{\rm max}+1) r_{\rm max}^2)$ entries, where $\nu_{\rm max} = \max_{k=0, \dots, d} \nu_k$ and $r_{\rm max} = \max_{k=1, \dots, d} r_k$.
Crucially, this is no longer exponential in $d$ if the ranks remain bounded.

Generally the tensor of function values at the collocation points \eqref{eq:collocation_tensor} is not of exact low rank, but
it has been observed that it can be approximated well with a low rank TT tensor for many (sufficiently smooth) functions appearing in practical applications \cite{osedelets11,osedelets10,dolgov2015simultaneous}.
Storing the function values at the collocation points as a TT tensor not only reduces the memory footprint as described above, but also allows for the interpolation to be done for each core separately.
In particular, assuming that $\Bv = \Bv^{(1)} \dots \Bv^{(d)}$, \eqref{eq:ip_tensorproduct} can be rewritten as
\begin{equation*}
  \tilde{g}(c_1, \dots, c_d)
  = \left( \sum_{j_1 = 0}^{\nu_1} \CL_{j_1}^1(c_1) \Bv^{(1)}(\colon, j_1, \colon) \right) \dots \left( \sum_{j_d = 0}^{\nu_d} \CL_{j_d}^d (c_d) \Bv^{(d)}(\colon, j_d, \colon) \right)
\end{equation*}
where each of the parentheses evaluates to a matrix of size $r_{k-1} \times r_k$ respectively, and the entire expression is a chain of matrix-matrix products. Since $r_0 = r_d = 1$, this evaluates to a scalar.

\subsubsection{Vector-valued tensor-trains}

Now consider $g = (g_i)_{i \in \{1, \dots, d_{\rm out}\}} \colon \R^{d_{\rm in}} \rightarrow \R^{d_{\rm out}}$.
As in \Cref{sec:vectorSmolyak}, we need to interpolate each of the $d_{\rm out}$ components.
This can be incorporated into the collocation scheme by adding another dimension to the tensor, such that
\begin{equation}\label{eq:tt_repr_tensor}
    \Bv(i, j_1, \dots j_d)
    = g_i(\xi_{j_1}^1, \dots, \xi_{j_{d_{\rm in}}}^{d_{\rm in}}).
\end{equation}
We again store this tensor in TT format $\Bv = \Bv^{(0)} \Bv^{(1)} \dots \Bv^{({d_{\rm in}})}$,
adding a $0$th core $\Bv^{(0)} \in \R^{r_{-1} \times d_{\rm out} \times r_0}$, where now $r_0 \in \N$ and $r_{-1} = 1$.
When evaluating the approximation $\tilde{g}$, interpolation in this new dimension is omitted, i.e.,
\begin{equation}\label{eq:TT_repr}
  \tilde{g}(c_1, \dots, c_d)
  = \Bv^{(0)} \left( \sum_{j_1 = 0}^{\nu_1} \CL_{j_1}^1 (c_1) \Bv^{(1)}(\colon, j_1, \colon) \right) \dots \left( \sum_{j_{d_{\rm in}} = 0}^{\nu_{d_{\rm in}}} \CL_{j_{d_{\rm in}}}^{d_{\rm in}} (c_{d_{\rm in}}) \Bv^{(d_{\rm in})}(\colon, j_{d_{\rm in}}, \colon) \right)
\end{equation}
The scheme in \eqref{eq:TT_repr} effectively is an approximation in the $r_0$-dimensional space spanned by the slices $\Bv^{(0)}[0,:,\alpha_0]$ of the $0$th TT core.
The coefficients with regards to these elements are given by the $r_0$ values appearing at the interface between the 0th and the 1st core, which are tensorized polynomials in $\bsc = (c_i)_{i \in \{ 1, \dots, d_{\rm in} \}}$.

\subsubsection{Operator surrogate}
The full surrogate of an operator $\CG$ is given by
\begin{equation}
  \TT = \CD \circ \tilde{g} \circ \CE
\end{equation}
with $\tilde{g}$ as in
\eqref{eq:TT_repr} and $\CE, \CD$ as introduced in \Cref{sec:enc_dec_architecture}.

\subsubsection{Evaluation of the TT surrogate}\label{section:tt_eval}

In \eqref{eq:TT_repr} we have already seen that the interpolation can be applied to each core separately.
This greatly reduces the computational cost of evaluating the collocation scheme compared to naive tensor interpolation as in
\eqref{eq:tt_repr_tensor}.
With the barycentric formulation introduced in \Cref{sec:surr_poly}, the interpolation in each core can be evaluated in $\cO(\nu_{\rm max})$
operations, followed by computing tensor contractions in $\cO(\nu_{\rm max} r_{\rm max}^2 + r_{\rm max}^3)$ operations.
In total it requires $\cO(d (\nu_{\rm max}+ r_{\rm max}) r_{\rm max}^2)$ operations to evaluate the TT representation.
Importantly, most FLOPs are spent on computing the tensor contractions, which are dense matrix-vector products.
Highly optimized BLAS routines for computing these operations are commonly available both for execution on CPU and GPU devices.

\subsubsection{Training: ALS-cross}\label{sec:als_cross}

Assuming that a given tensor $\Bv$ admits approximation using a low rank TT tensor, the next question becomes how to find this TT tensor.
A quasi-optimal low rank approximation can be obtained by computing a sequence of SVD of so-called unfolding matrices \cite{osedelets11}, requiring knowledge of the full tensor,
which is infeasible for large dimension $d$.
This leads to cheaper decomposition algorithms such as TT cross approximation \citep{osedelets10}.
However, this method is iterative and requires $\cO(d \nu_{\rm max} r_{\rm max}^2)$ evaluations of the approximated function $\tilde g$ for every pass.
For expensive forward operators, such as those given by PDE solution operators, this becomes intractable even at moderate ranks.
Thus, more specialized methods are required for such operators.

One method is the ALS cross algorithm \citep{dolgov18}, which takes advantage of the additional structure provided by the PDE problem.
In particular, it requires the truncated coefficient map $ g_{d_{\rm in}}^{d_{\rm out}}$ as in \eqref{eq:gtruncref} to be given by the solution to a linear equation, i.e.,
\begin{equation*}
    A(x) g(\CE (x)) = b(x)
\end{equation*} with $A(x) \in \R^{d_{\rm out} \times d_{\rm out}}$ and $b(x) \in \R^{d_{\rm out}}$, both depending on the input $x\in\CX$.
Recall that the TT surrogate is a collocation method.
Choosing collocation points $(\xi_{j_k}^k)_{j_k \in \{0, \dots, \nu_k\}} \subset \R$ for each $k \in \{1, \dots, d_{\rm in}\}$, we write in abuse of notation for any $v$ dependent on $\CE_{d_{\rm in}}(x) \in \R^{d_{\rm in}}$
\begin{equation*}
    v(j_1, \dots, j_{d_{\rm in}}) = v((\xi_{j_1}^1, \dots, \xi_{j_{d_{\rm in}}}^{d_{\rm in}}))
    \qquad \forall k \in \{1, \dots, {d_{\rm in}}\}: j_k \in \{0, \dots, \nu_k \}.
\end{equation*}
The method then solves the parametric linear equations
\begin{equation}\label{eq:linear_system_coll}
    A(j_1, \dots, j_{d_{\rm in}}) u(j_1, \dots, j_{d_{\rm in}}) = b(j_1, \dots, j_{d_{\rm in}})
    \qquad \forall k \in \{1, \dots, {d_{\rm in}}\}: j_k \in \{0, \dots, \nu_k \}
\end{equation}
on the grid of collocation points, where $u(j_1, \dots, j_{d_{\rm in}}) \in \R^{d_{\rm out}}$.
The parametric solution tensor $\Bu$ of \eqref{eq:linear_system_coll} then gives the surrogate map $\tilde g \colon \R^{d_{\rm in}} \to \R^{d_{\rm out}}$
in the form \eqref{eq:TT_repr}.
By working with TT approximations of the components of the equation, this remains tractable even for higher values of $d$.
A crucial requirement for the applicability of the algorithm is the availability of TT approximations $\BA$ and $\Bb$ of $A(j_1, \dots, j_{d_{\rm in}})$ and $b(j_1, \dots, j_{d_{\rm in}})$.

Application to parametrized linear PDEs is done by solving the associated linear equation given by discretization.
In general, obtaining the TT approximations $\BA$ and $\Bb$ using, e.g., cross approximation directly is not feasible, as evaluating entries of $A(x)$ and $b(x)$ requires expensive assembly of linear system associated with the parameter $x$.
However, the parameter dependency of $A$ and $b$ usually stems from some (physical) scalar parameter fields $K_i = K_i(x)$
appearing in the weak formulation of the PDE, which are given by some straightforward closed-form expression.
Think of, e.g., diffusivity in the heat equation or Lam\'e parameters for elasticity.
After choosing a discretization (e.g.\  using the FE basis), TT cross approximation can be used to compute a TT approximation
$\BK_i$ of $K_i$ at the collocation points.
Assume that
\begin{equation*}
    A(j_1, \dots, j_{d_{\rm in}})
    = \sum_i \underbrace{L_i^A (\BK_i ((j_1, \dots, j_{d_{\rm in}})))}_{=: A_i(j_1, \dots, j_{d_{\rm in}})},
    \qquad
    b(j_1, \dots, j_{d_{\rm in}})
    = \sum_i \underbrace{L_i^b (\BK_i((j_1, \dots, j_{d_{\rm in}})))}_{=: b_i(j_1, \dots, j_{d_{\rm in}})}
\end{equation*}
where $L_i^A,  L_i^b$ are linear maps mapping the physical parameters $K_i$ to the respective component of $A_i \in \R^{d_{\rm out} \times d_{\rm out}}$ and $b_i \in \R^{d_{\rm out}}$.
Then the TT approximations $\BA$ and $\Bb$ can be obtained by assembling a small number of systems of discretized equations.
More precisely, evaluations of $A_i, b_i$ are required, with the number of evaluations required depending on the
ranks of the $\BK_i$.
In practice, this requires splitting up the left-hand side and the right-hand side of the discretized PDE into a sum of components linearly dependent
on one parameter.
This is a mostly straightforward modification of most numerical PDE codes, but it also means that the ALS cross is not a black box algorithm.

While a priori dimension reduction on the output space using orthogonal projections like PCA is possible with the ALS cross algorithm, it is originally designed to be applied directly to the PDE discretization.
In the case of e.g., FEM, the decoder only consists of the coefficient-to-function map of the FE discretization, while a suitable reduction from the potentially very high-dimensional FE space to a $r_0$-dimensional approximation subspace is computed by the ALS cross algorithm.
Note that this corresponds to an adaptive decoder dimension $d_{\rm out} = r_0$.

\section{Approximation theory and convergence rates} \label{sec:approximation_theory}

In this section, we briefly recapitulate some established approximation results for the operator surrogates we consider in this work. Both the theoretical convergence rates as well as the practical performance typically depend on smoothness properties of the target operator and the input distribution.
Common assumptions on the operator $\CG$ and the input distribution are of the following type.

\begin{assumption}[{\citep[Assumption~1]{herrmann2024neural}}]\label{asm:ops:holomorphy}
There exist $s>1, t>0, M<\infty$ and an open set $O_{\C} \subseteq \CX_\C$ containing $K^s$ such that $\sup _{x\in O_\C}\|\CG(x)\|_{\CY_\C^t} \leq M$ and $\CG: O_\C \rightarrow \CY_\C$ is holomorphic.
\end{assumption}

Under this assumption, reduced-basis neural operator surrogates (irrespective of whether $L^2_\mu$ or $H^1_\mu$ is used) admit the following expression rates as the network size increases.

\begin{theorem}[{\citep[Theorem~2]{herrmann2024neural}}] \label{thm:ops:rbno}
Let \Cref{asm:ops:holomorphy} be satisfied with $s>1, t>0$. Fix $\delta>0$ (arbitrarily small). Then, there exists a constant $C>0$ such that for every $N \in \N$, there exists a ReLU NN $\tilde g$ of size $N$ such that
$$
\sup_{x \in K^{s}}
\norm[\CY]{\CG(x)-\CD \circ\tilde g \circ \CE(x)}
\leq C N^{-\min \{s-1, t\}+\delta}.
$$
\end{theorem}

We emphasize that the above result expresses convergence with increasing \textit{parametric} complexity, measured by the number of trainable parameters $N$. For results on convergence with increasing \textit{data} complexity, measured as the number of training data $n$, see \citep{reinhardt2024statistical}. Note that the setting of that work differs slightly in that it considers empirical risk minimization under noisy data.

FNOs are universal approximators, and have also been shown to admit algebraic convergence rates for certain PDEs \citep{kovachki2021universal}. This is achieved by showing that the FNO can emulate a numerical algorithm to solve the method. We next recall a statement from \citep{kovachki2021universal} establishing convergence for the Darcy flow on the $d$-dimensional torus $\bbT^d$. The statement is shown for an FNO variant denoted as $\Psi$-FNO, which replaces the exact Fourier transform in each layer by a discrete Fourier transform, making the spectral operations finitely parametrizable and computable, while the architecture still acts as an operator between function spaces in the spatial variable.

\begin{theorem}[{\citep[Theorem~26]{kovachki2021universal}}]\label{thm:fno}
  Let $d$, $k \in \mathbb{N}$ and $\lambda\in(0,1)$, and let $\mathcal{A}_{\lambda}^{s}(\mathbb{T}^{d})=\set{a\in H^{s}(\mathbb{T}^{d})}{\norm[H^s(\bbT^d)]{a}\le\lambda^{-1},~{\rm essinf}_{z\in\bbT^d}{a(z)}\ge\lambda}$ with smoothness $s \ge d/2 + k + \delta$ for some $\delta \in (0,1)$. Let $\mathcal{G}: \mathcal{A}_{\lambda}^{s}(\mathbb{T}^{d}) \rightarrow H^{1}(\mathbb{T}^{d})$ be the solution operator $a\mapsto u$ corresponding to the PDE
  \begin{equation}\label{eq:ellpde}
    \begin{aligned}
  -\nabla ( a\cdot\nabla u)&=f &&\text{in }\bbT^d\\
      \int_{\bbT^d}u&=0,
    \end{aligned}
\end{equation}
for fixed $f\in H^{k-1}(\bbT^d)$ with zero mean. Then there exists a constant $C > 0$ such that for any $N \in \mathbb{N}$, there exists a $\Psi$-FNO $\tilde{g}$ satisfying $$\sup_{a \in \mathcal{A}_{\lambda}^{s}} \|\mathcal{G}(a) - \tilde{g}(a)\|_{H^{1}(\mathbb{T}^{d})} \le C N^{-k}$$with the number of trainable network parameters of the surrogate $\tilde g$ bounded by $N^{d} \log(N)$.
\end{theorem}

\begin{remark}
  For the PDE \eqref{eq:ellpde}, the convergence rate of FNO provided by Theorem \ref{thm:fno} is typically better than the one implied by Theorem \ref{thm:ops:rbno} for Fourier-type encoders/decoders, cf.~\cite[Section 7.1]{herrmann2024neural}. Nonetheless, Theorem \ref{thm:ops:rbno} has two advantages: First, it treats the whole class of holomorphic operators, which is known to include a wide range of parametric PDEs, e.g., \cite{CoDe}. Second, the representation of the output via frames allows to treat arbitrary domains both in theory and practice, cf.~\cite[Section 7.2]{herrmann2024neural}; the FNO's structure, which strongly relies on fast implementations of the Fourier transform, typically requires discretization on uniform grids and rectangular domains.
\end{remark}

An analysis of the discretization error occurring in practice when using finite-element representations of the operator input and output can be found in \citep{lanthaler2024discretization}. For an error analysis focusing on convergence with increasing data complexity, see \citep{kovachki2024data}.

In contrast to neural operator surrogates, for interpolation-based operator surrogates $\Smolyak$, parameter and data complexity are equivalent and given as the number of interpolation nodes. Under \Cref{asm:ops:holomorphy}, $\Smolyak$ admits the following convergence rates for the worst-case error as the number of interpolation nodes increases.

\begin{theorem}[{\citep[Theorem~4]{herrmann2024neural}}]\label{thm:ops:interpolation}
Let \Cref{asm:ops:holomorphy} be satisfied with $s>1, t>0$. Fix $\delta>0$ (arbitrarily small). Then, there is a constant $C>0$ such that for every $n \in \N$, there exists a multivariate polynomial $\tilde g$ such that
$$
\sup _{x \in K^{s}}
\norm[\CY]{\CG(x)-\CD \circ \tilde g \circ \CE (x)}
\leq C n^{-\min \{s-1, t\}+\delta}.
$$
Furthermore, $\tilde g$ belongs to an $n$-dimensional space of multivariate polynomials. Its components are interpolation polynomials, whose computation requires the evaluation of $\left\langle \CG(x), \psi_j\right\rangle_\CY$ at at most $n$ tuples $(x, j) \in K^s \times \N$.
\end{theorem}

For $\TT$, the following results on convergence with increasing parametric complexity were shown in \cite{ttexpressionrate2026}.

\begin{theorem}[{\citep{ttexpressionrate2026}}]
  \label{thm:tt_expression_rate}
  Let \Cref{asm:ops:holomorphy} hold.
  Fix $\delta>0, r>0$ and set $\tilde{s} = s - \delta$.
  Then there exists a constant $C > 0$ such that for every $N\in\N$ there exists a TT representation
  $\tilde{g}$ of size $O(N)$ such that
  \begin{equation*}
    \sup_{x\in K^{s}}\norm[\CY]{\CG(x)- (\CD \circ \tilde{g} \circ \CE)(x)}
    \le C
    \begin{cases}
      N^{-(\tilde{s}-1) / 3 + \delta} & \tilde{s}-1 \leq 2 t \\
      N^{- \frac{t (\tilde{s}-1)}{t + \tilde{s} - 1} + \delta} & \tilde{s} - 1 \geq 2 t.
    \end{cases}
  \end{equation*}
\end{theorem}

Under slightly modified assumptions, the same rates can be shown for TT surrogates without the additional dimension reduction in the decoder, cf.\ \cite{ttexpressionrate2026}.

\section{Empirical benchmarking} \label{sec:numerical_results}

In this section, we compare the methods described in \Cref{sec:surr_neural,sec:surr_poly}: For neural network-based surrogates, we consider the reduced-basis neural operator with training in $L^2_\mu$ and $H^1_\mu$ ($\LTwoRBNO$ and $\HOneRBNO$, respectively), and the Fourier neural operator ($\FNO$) with training in $L^2_\mu$. For polynomial surrogates, we consider the reduced-basis sparse-grid surrogate ($\Smolyak$) and the TT surrogate ($\TT$). The operator surrogates are applied to a linear parametric diffusion equation and a nonlinear parametric hyperelasticity problem. Key points of comparison are the cost-accuracy tradeoffs with regard to the offline sample cost in \Cref{sec:exp_error_training_data} and the online evaluation cost in \Cref{sec:exp_error_eval_cost}.

The experiments presented here are based on \citep[Chapter~3]{westermann2026polynomial} and extend this work as follows. First, the TT-based surrogate was added to the range of surrogates studied. The set of hyperparameters of the neural operator surrogates has been extended, and now includes widths 400 and 800 as well as depth 7 for $\LTwoRBNO$ and $\HOneRBNO$, as well as $\FNO$ setups with 64 channels and 16 and 32 Fourier modes.
Finally, the maximal number of encoder and decoder dimensions has been increased from 400 to 800.

\subsection{Setting} \label{sec:exp_probs}

Our test problems are based on solution operators for parametric PDEs, i.e., the parameter-to-solution map $\CG:\CX \to \CY$ with $\CR(x, \CG(x))= 0$ for a differential operator $\CR$.
In particular, we consider a linear elliptic and a nonlinear PDE. In both applications, the spatial domain is the unit square $[0,1]^2$, and we discretize the parameter field and the solution on a uniform $64\times64$ grid using continuous Q1 finite elements.

\subsubsection{Parameter field} \label{sec:parameter_field}

\begin{figure}[H]
  \centering
  \includegraphics[width=\textwidth]{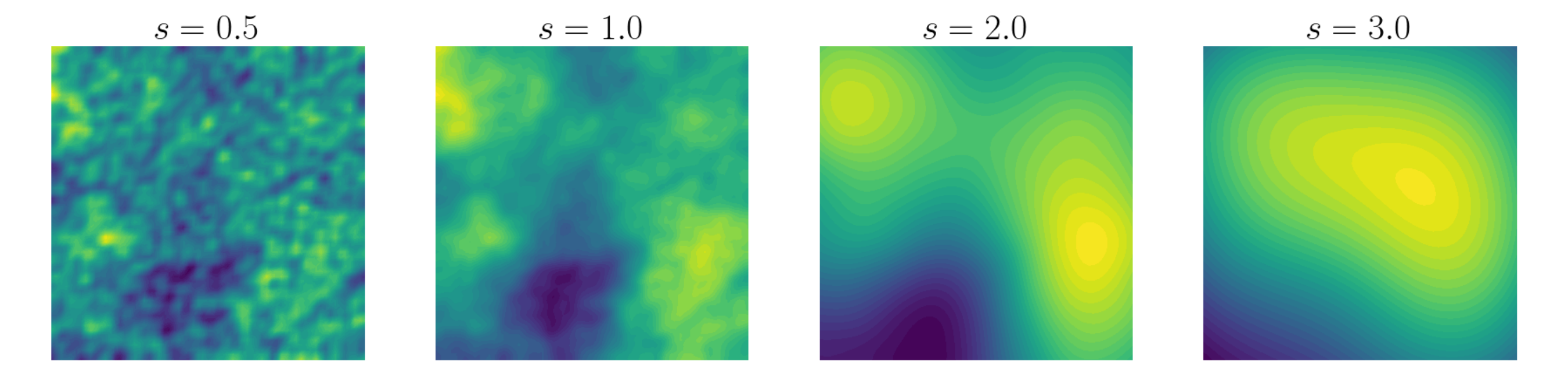}
  \caption{
  Input samples constructed as in \eqref{eq:samples} using a fixed coefficient vector $\bsc$ and varying $s$ values. As $s$ decreases, the correlation length becomes smaller.}
  \label{fig:samples}
\end{figure}

Both PDEs use as parametric input a scalar field $x \in \CX = L^2(\Omega) = {\rm span} \{\psi_j\}_{j\in\N}$. For our experiments, we choose $\psi_j$ to be the eigenfunctions of a Mat\'ern covariance of the form
\begin{equation*}
    (\gamma \text{Id} - \delta\Delta)^{-2} \psi_j = \mu_j \psi_j
\end{equation*}
with Robin boundary conditions. The eigenfunctions are computed using generalized eigenvalue solvers implemented in \cite{VillaPetraGhattas18,oleary2021hippyflow}, for more details see \cite{villa2024note}. The parameters $\gamma$ and $\delta$ control the marginal variance and correlation length for the random parameters.
In order to control the smoothness in the sense of \eqref{eq:Ks}, we compute the $\psi_j$ using fixed $\gamma=0.1$ and $\delta = 0.5$, and express inputs as
\begin{equation}\label{eq:samples}
    x(\bsc) := \sum_{j=1}^{d_{\rm true}} c_j  j^{-s} \psi_j,
    \qquad \bsc = (c_j)_{j= 1}^{d_{\rm true}} \in [-1,1]^{d_{\rm true}}
\end{equation}
with $d_{\rm true} = 1000$.
Here, the weights $j^{-s}$ correspond to $\lambda_j = j^{-1}$ in the notation of \eqref{eq:Ks}, so that the inputs \eqref{eq:samples} lie in $K^s$. The Mat\'ern eigenvalues $\mu_j$ from the eigenproblem above are not used. Due to $d_{\rm true}<\infty$, it holds $x(\bsc)\in K^{\tilde s}$ for any $\tilde s\ge 0$; however, as $d\to\infty$, we get for $s\ge 1/2$ that $x_d(\bsx)=\sum_{j=1}^d c_j j^{-s}\psi_j\in K^{s-1/2}$, with a $d$-independent bound on $\|x_d(\bsx)\|_{H^{s-1/2}}$.
\Cref{fig:samples} visualizes how $s$ controls the smoothness for a fixed coefficient $\bsc$.

\subsubsection{Test problem 1: Diffusion equation} \label{sec:poisson}

\begin{figure}[H]
  \centering
  \includegraphics[width=.7\textwidth]{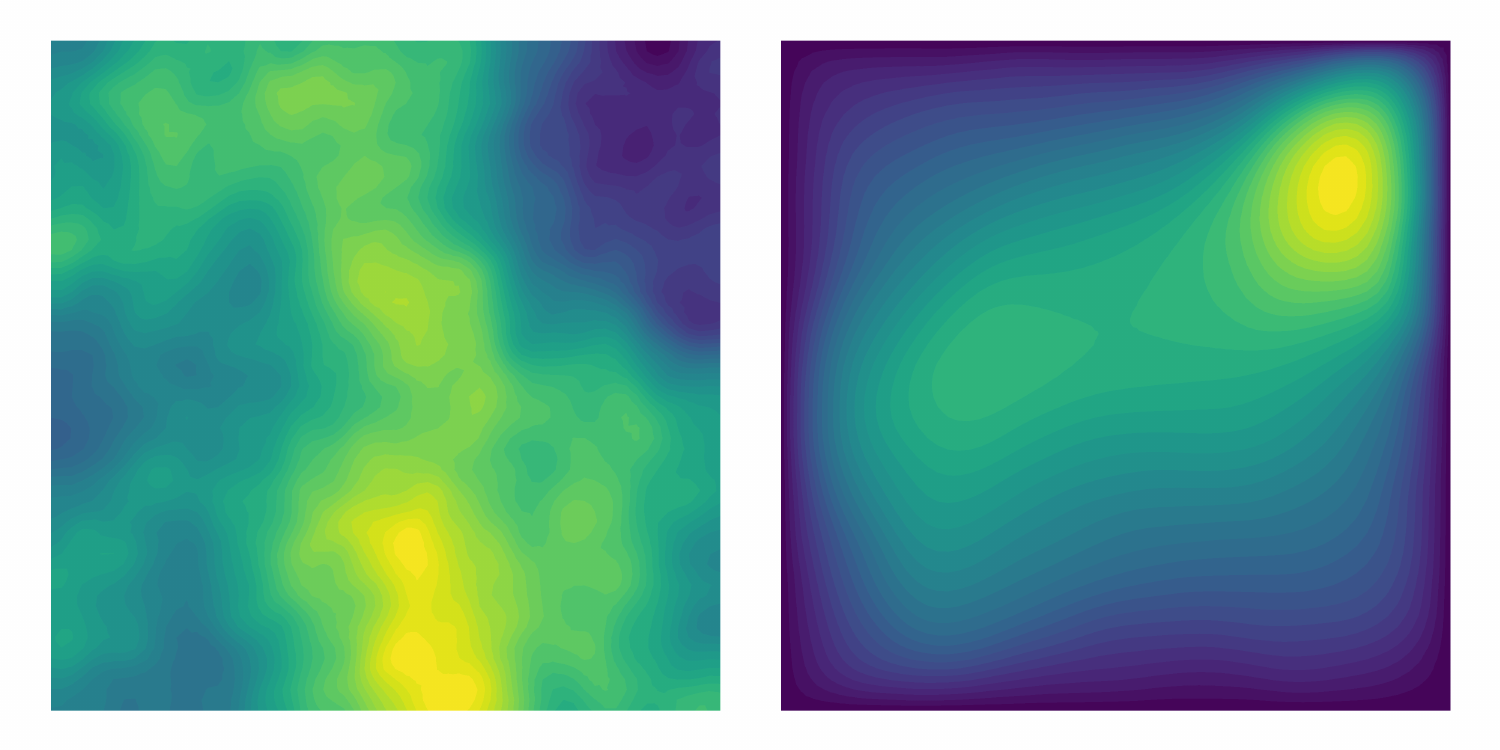}
  \caption{A random parameter field $x(\bsxi)$ (left) and corresponding FE solution $y(\bsxi)$ of the linear elliptic PDE \eqref{eq:elliptic} (right).}
  \label{fig:exemplary_poisson}
\end{figure}

We consider a second-order elliptic PDE in divergence form given by
\begin{equation}\label{eq:elliptic}
\begin{aligned}
    - \nabla \cdot (e^{x(\bsc)} \nabla y) &= 1 \quad  \text{ in } \Omega := [0,1]^2 \\
    y &= 0 \quad \text{ on } \partial \Omega,
\end{aligned}
\end{equation}
with log-permeability field $x$ as in \Cref{sec:parameter_field}.
The corresponding input and output spaces are $\CX = L^2(\Omega)$ and $\CY = H_0^1(\Omega)$.
This problem is linear and exhibits smooth solutions that can be well approximated using a linear decoder as described in \eqref{eq:dec}. See also \Cref{fig:exemplary_poisson} for an example of a data pair.

\subsubsection{Test problem 2: Hyperelasticity}\label{section:hyperelasticity}

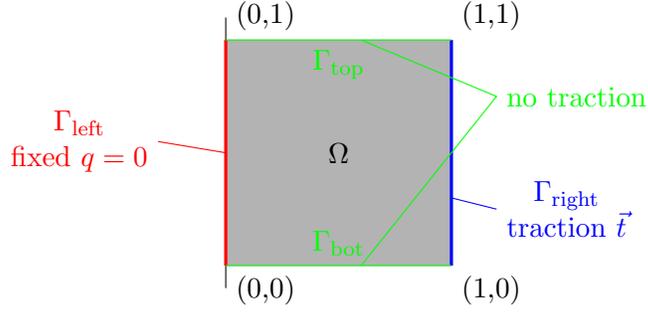
\begin{figure}
    \centering
\begin{tikzpicture}[scale=3]
    \coordinate (bl) at (0,0);
    \coordinate (br) at (1,0);
    \coordinate (tr) at (1, 1);
    \coordinate (tl) at (0, 1);

    \filldraw[color=black!30]
    (bl) --
    (br) --
    (tr) --
    (tl) -- cycle;
    \draw (.5,.5) node {$\Omega$};

    \filldraw[draw opacity=0, pattern={north east lines}]
    (0, -.1) --  (0, 1.1) -- (-.2,1.1) --   (-.2,-.1) --cycle;

    \draw (0, -.1) --  (0, 1.1);
    \draw[very thick,color=red] (0,0) --  (0,1);
    \draw[color=red] (0,.5) -- +(-.3, .05) node[left,align=center] {$\Gamma_{\textrm{left}}$ \\ fixed $q=0$};
    \draw[very thick,color=blue] (1,0) --  (1,1);
    \draw[color=blue] (1,.3) -- +(.2,-.05) node[right, align=center] {$\Gamma_{\textrm{right}}$ \\ traction $\vec{t}$};

    \draw[color=green] (.6,0) -- (1.2, .75) node[right] {no traction} -- (.6,1);

    \draw[color=green] (bl) -- node[above, opacity=1] {$\Gamma_{\textrm{bot}}$} (br);
    \draw[color=green] (tr) -- node[below] {$\Gamma_{\textrm{top}}$} (tl);

    \draw (bl) node[anchor=north west] {(0,0)};
    \draw (tl) node[anchor=south west] {(0,1)};
    \draw (br) node[anchor=north west] {(1,0)};
    \draw (tr) node[anchor=south west] {(1,1)};
\end{tikzpicture}
    \caption{2D deformation problem}
    \label{fig:elasticity_problem}
\end{figure}

In order to include a nonlinear PDE into our study, we consider the deformation of a hyperelastic material under stress using the problem setup introduced in \citep{cao2023residual} and subsequently employed in \citep{o2024derivative}.
Here, a two-dimensional square material, fixed on one side of its domain, is subject to a traction force that deforms the material (cf.\ \Cref{fig:elasticity_problem}).
Denoting with $y : \Omega = [0,1]^2 \to \R^2$ the unknown displacement field, the new coordinates of a material point $\bsxi \in \Omega$
after deformation are given by $\chi(\bsxi) = \bsxi + y(\bsxi)$.
Parametric input to this problem is a spatially varying stiffness modulus (also termed Young's modulus)
$ E(x, \bsxi) = 1 + \exp (x(\bsxi))$, where - generalizing the setup from \citep{cao2023residual,o2024derivative} - we choose $p$ as in \Cref{sec:parameter_field}.
In the following, we will omit the parametric input variable for the sake of brevity.
The displacement field $y$ then solves the steady state differential equation
\begin{equation}\label{eq:hyperelasticity_strong}
  \nabla \cdot(F (y) S (x))=0 \text { in } \Omega
\end{equation}
with boundary conditions
\begin{equation*}
  y=0 \text { on } \Gamma_{\text {left }}\qquad\text{and}\qquad
  F S \cdot n=
                          \begin{cases}
  0 &\text{on } \Gamma_{\text {top }} \cup \Gamma_{\text {bot }} \\
  t &\text{on } \Gamma_{\text {right}}
                            \end{cases}
\end{equation*}
where $F$ is the deformation gradient $F(\bsxi) = \nabla \chi(\bsxi) = I + \nabla y(\bsxi)$ and $S$ is the second Piola-Kirchhoff stress tensor, characterizing the stress in the reference configuration. With the Cauchy-Green stress tensor $C=F^{\top} F$,
$$
  S(\bsxi, C) = 2 \frac{\partial W(\bsxi, C)}{\partial C}
$$
where $W$ is the strain energy density of a neo-Hookean material given as
\begin{equation}\label{eq:strainenergy}
  W(\bsxi, C)=\frac{\mu(\bsxi)}{2} (\operatorname{tr}(C)-3)+\frac{\lambda(\bsxi)}{2} \ln (J)^2-\mu(\bsxi) \ln (J).
\end{equation}
Here, $J=\operatorname{det}(F) = \sqrt{\operatorname{det}(C)}$ and $\lambda, \mu$ are the Lam\'e parameters
\begin{equation*}\label{eq:lame_parameters}
    \lambda(\bsxi) = E(\bsxi) \frac{ \nu_P}{\left(1+\nu_P\right)\left(1-2 \nu_P\right)}, \quad
  \mu(\bsxi) = E(\bsxi) \frac{1}{2\left(1+\nu_P\right)}.
\end{equation*}
Poisson's ratio $\nu_P$ is assumed to be $0.4$. Finally, the traction vector $t$ acting on $\Gamma_{\text{right}}$ is given as a combination of a Gaussian compressive component and a linear shear contribution,
\[ t(\bsxi) :=
\begin{pmatrix}
0.06 \exp \left(-0.25\left|\xi_2-0.5\right|^2\right)\\
0.03\left(1+0.1 \xi_2\right)
\end{pmatrix}, \qquad \bsxi = (\xi_1, \xi_2).
\]
Here, the corresponding input and output spaces are $\CX = L^2(\Omega)$ and $\CY = H^1(\Omega) \times H^1(\Omega)$.

The nonlinear strain energy density in \eqref{eq:strainenergy} leads to a nonlinear PDE, which is solved using a Newton iteration for the reference solution.
Due to this nonlinearity, the ALS cross algorithm from \Cref{sec:als_cross} can not be applied, and we thus omit TT surrogates from the experiments with hyperelasticity.

\begin{figure}[H]
  \centering
  \includegraphics[width=.7\textwidth]{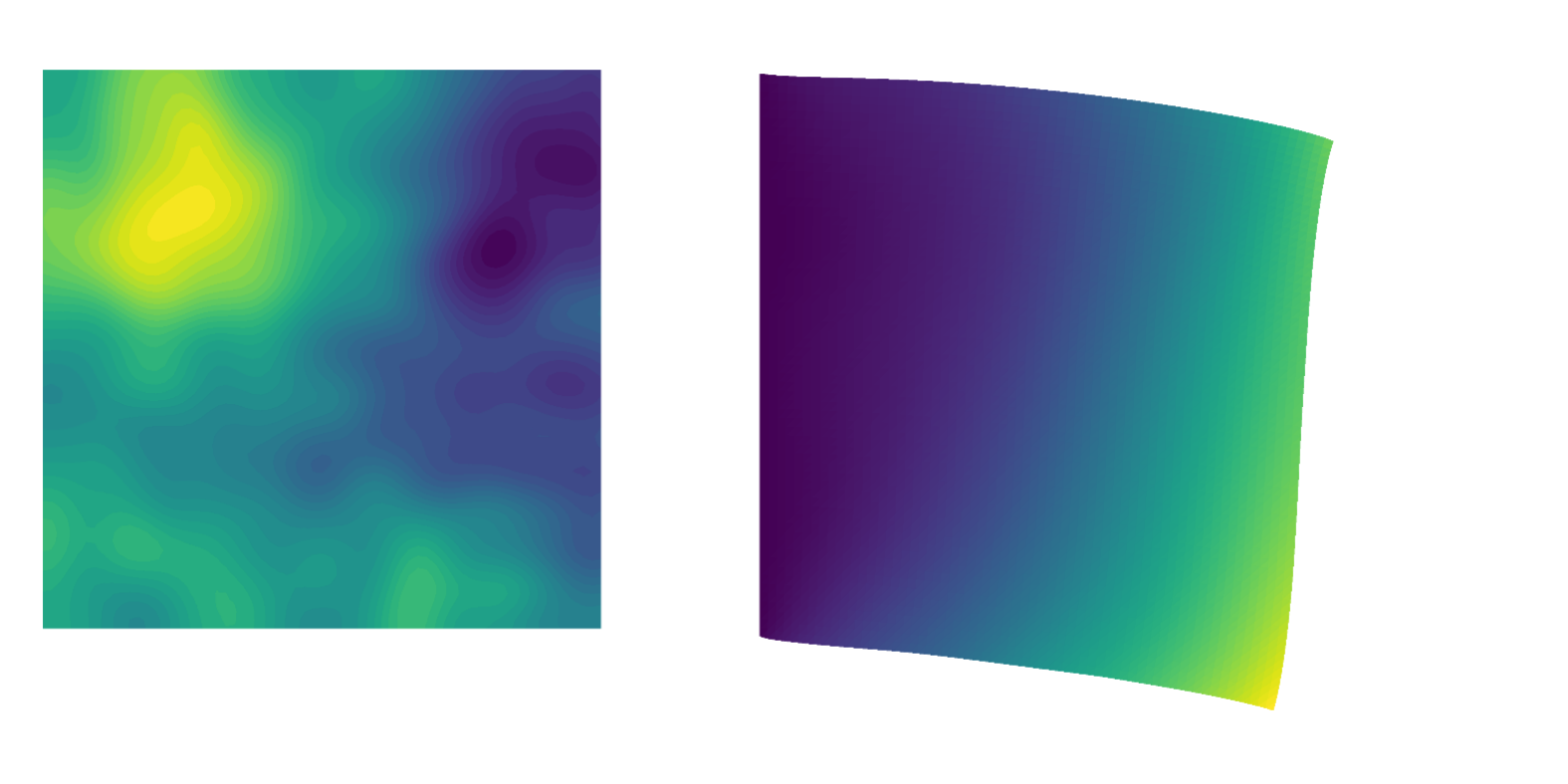}
  \caption{A random parameter field $x(\bsxi)$ (left)
   and corresponding FE solution $y(\bsxi)$ of the hyperelasticity problem (right).}
  \label{fig:exemplary_hyperelasticity}
\end{figure}

\subsubsection{Measurement of surrogate accuracy and cost}\label{sec:measuring}

\paragraph{$L^2_\mu$-error.}

A key quantity to measure the accuracy of a PDE surrogate $\tilde{\CG}$ is the relative error in the $L^2(\CX, \mu; \CY)$ sense, i.e.\ the relative root mean square (RMS) error in $\CY$, cf.\ \eqref{eq:L2obj}.
We estimate this error via Monte-Carlo approximation as
\begin{equation}\label{eq:L2mu}
    \varepsilon_{L^2_\mu}(\tilde{\CG}) := \sqrt{\frac{\sum_{k=1}^K  \normOut{y^{(k)} - \tilde{\CG} (x^{(k)})}^2}{\sum_{k=1}^K \normOut{y^{(k)}}^2}}
\end{equation}
given $K=250$ parameter samples $(x^{(k)})_{k=1}^K$ and corresponding solutions $y^{(k)} = \CG(x^{(k)})$.
In our numerical experiments, we neglect the finite element discretization error and consider the discrete solution operator $\CG$ as the exact operator to be learned.
As this applies to all surrogate methods equally, the relative performance stays unchanged.
However, it should be kept in mind that the errors relative to the true PDE solution might be larger than the numbers shown here.

\paragraph{$H^1_\mu$-error.}

As discussed in \Cref{sec:training_L2_H1}, certain applications also require gradient data with regard to the parameters.
This error can be quantified in a $H^1(\CX, \mu; \CY)$ sense.
Specifically, we consider the relative RMS error in the ${\rm HS}(\CX^s, \CY)$ seminorm.
In line with \eqref{eq:H1objtildeg} for $\tilde{s} = s$, this can be estimated as follows

\begin{equation}\label{eq:H1mu}
    \varepsilon_{H^1_\mu}(\tilde{\CG})
    := \sqrt{
            \frac{
            \sum_{k=1}^K  \bnorm*{\left(\nabla g_{d_{\rm in}}^{d_{\rm out}} (x^{(k)}) - \nabla \tilde g (x^{(k)})\right) \cdot \operatorname{diag}(\Blambda^s)}_{\rm F}^2
            }
            {\sum_{k=1}^K \bnorm*{\left( \nabla g_{d_{\rm in}}^{d_{\rm out}} (x^{(k)}) \right) \cdot \operatorname{diag}(\Blambda^s)}_{\rm F}^2}
        },
\end{equation}
with $g_{d_{\rm in}}^{d_{\rm out}}$ as defined in \eqref{eq:gtruncref}.

\paragraph{Number of training data points.}

We denote the number of training data points by $n$.
For PDE applications, where acquiring training data pairs using a deterministic solver is potentially expensive, this is often a dominant contributor to the \emph{offline} cost associated with a surrogate, i.e., the one-time cost to build the surrogate.
For polynomial surrogates, the latter corresponds to the number of interpolation nodes, and for TT surrogates, we count the number of deterministic solves the ALS cross algorithm requires.
However, to avoid redundant terminology, we often refer to all of these as the number of training data.

While this metric is not as holistic as, e.g., simply measuring the runtime of the offline computation, it is largely independent from implementation specifics and hardware.
Thus, it provides a fairer comparison of the underlying methods themselves under the assumption of data acquisition dominating the cost.

\paragraph{Evaluation time.}

Operator surrogates aim to reduce the evaluation cost in many-query applications.
Thus, low cost to evaluate new samples at a given accuracy level once the surrogate is built is crucial.
These costs are commonly referred to as \emph{online} costs.
We will denote the amortized walltime per sample by $t_E$.

\paragraph{Training time.}

We also measure the walltime of the offline training/computation, excluding the time to compute samples, and refer to this as $t_T$.
It is important to note that this quantity can be quite sensitive to hyperparameter choice, software implementation, and computing hardware.

\subsubsection{Software choices}

The neural networks used for $\LTwoRBNO$ and $\HOneRBNO$ are implemented via standard functionality in PyTorch \cite{paszke2019pytorch}, and JAX \cite{bradbury2018jax}. For $\FNO$, we use the implementation provided by \citep{huang2025an}. For $\Smolyak$, the interpolant is implemented in Smolyax \citep{westermann2025smolyax}, a GPU-capable implementation of sparse-grid interpolation in JAX.
Lastly, the ALS cross algorithm used for offline computation of $\TT$ is currently only implemented on top of NumPy \cite{harris2020array} and thus limited to CPU devices.
The online evaluation can be done using callbacks to precompiled binaries utilizing a suitable BLAS backend (e.g., OpenBLAS or Intel MKL), as well as within both PyTorch and JAX. For our specific benchmarks run on GPU machines, the JAX implementation was used. The PDE implementations were implemented in FEniCS \cite{logg2012automated} using hippylib \cite{VillaPetraGhattas18} for modeling random fields and derivative computations as well as hippyflow \cite{oleary2021hippyflow} for reduced basis computations and training data generation.

\subsection{Hyperparameters} \label{sec:exp_hyperpars}

Before turning to a comprehensive comparison of surrogate performance in terms of accuracy and computational cost, we first study how surrogate design choices affect surrogate accuracy. This includes network architectures and activation functions for neural operators (\Cref{sec:hyperpars_no}), the choice of polynomial spaces for the sparse-grid surrogate (\Cref{sec:hyperpars_ip}), number of dimensions and orthogonality of the en- and decoder (\Cref{ssec:enc_dec_setup}), as well as initialization, polynomial degrees and ALS iteration count for the tensor-train surrogate (\Cref{ssec:tt_setup}). We emphasize that it is generally impossible to search the space of hyperparameters exhaustively; our main aim is to motivate hyperparameter choices and to establish an understanding of how they affect surrogate accuracy. For brevity, we report only the results for the diffusion problem since the results for the hyperelasticity problem are qualitatively similar.

\subsubsection{Neural operator architectures} \label{sec:hyperpars_no}

For training the neural operators, we apply the ADAM optimizer \cite{kingma2017adam} for 2000 epochs with a batch size $n_{\rm batch} = 32$ and a piecewise constant learning rate schedule. Further, $5 \%$ of the training data is used for model validation.

\begin{figure}
    \centering
    \includegraphics[width=\linewidth]{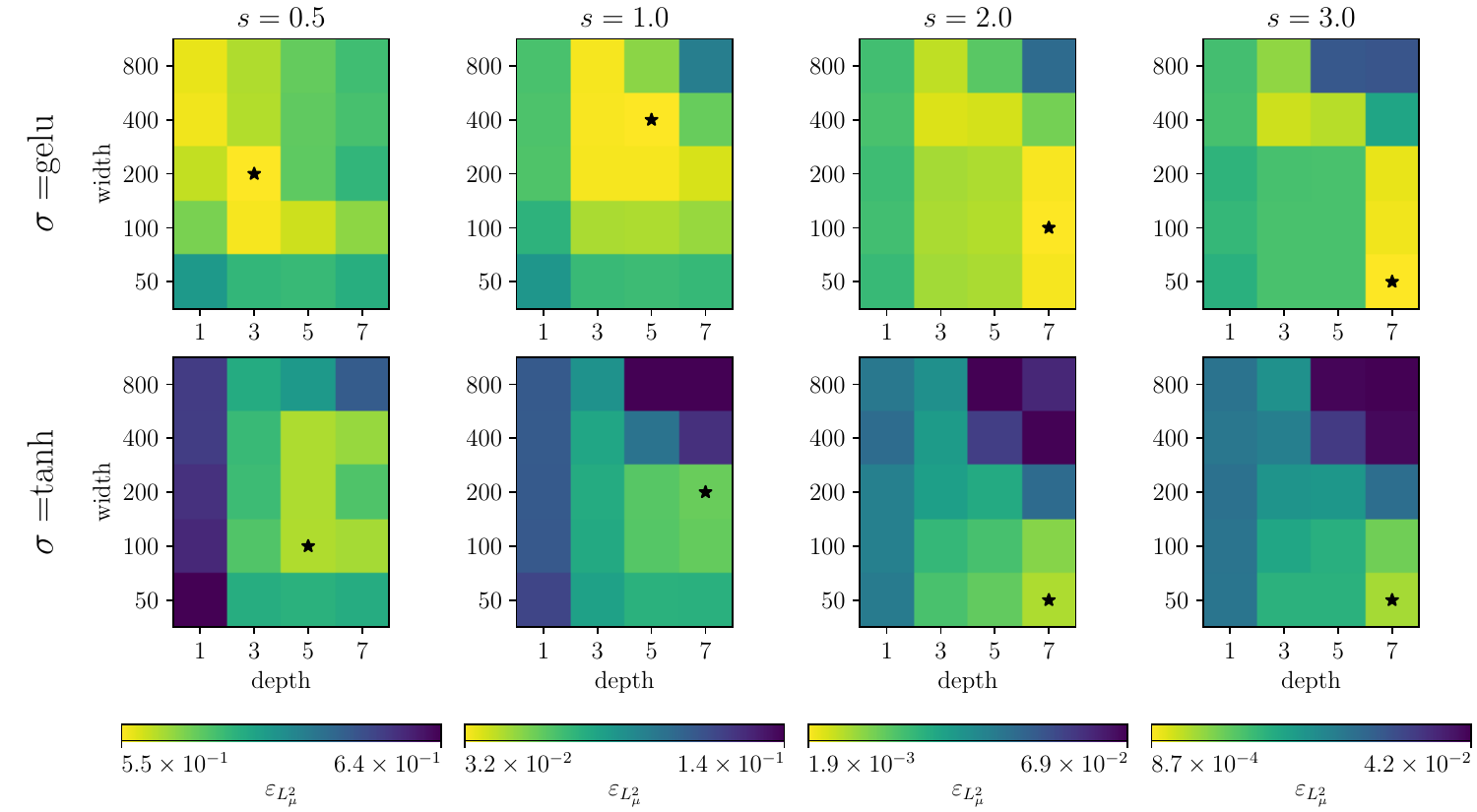}
    \caption{\small $\varepsilon_{L^2_\mu}(\LTwoRBNO)$ for the diffusion problem \eqref{eq:elliptic} over network depth and width for different input smoothness $s$. Top row: GELU activation function; bottom row: tanh activation function. Coefficient-space dimensions fixed at $d_{\rm in}=d_{\rm out}=200$, training set size $n=10^4$.}
    \label{fig:ops_hyperpars_rbno}
\end{figure}

\Cref{fig:ops_hyperpars_rbno} shows $\LTwoRBNO$ accuracies across different input smoothness scales, network architectures, and for both GELU and tanh activation functions. The results clearly show that using the GELU activation function is advantageous compared to using the tanh activation function across all settings. For the subsequent analysis, we therefore fix the activation function to GELU, \citep{hendrycks2023gelu}.

\begin{figure}
    \centering
    \includegraphics[width=\linewidth]{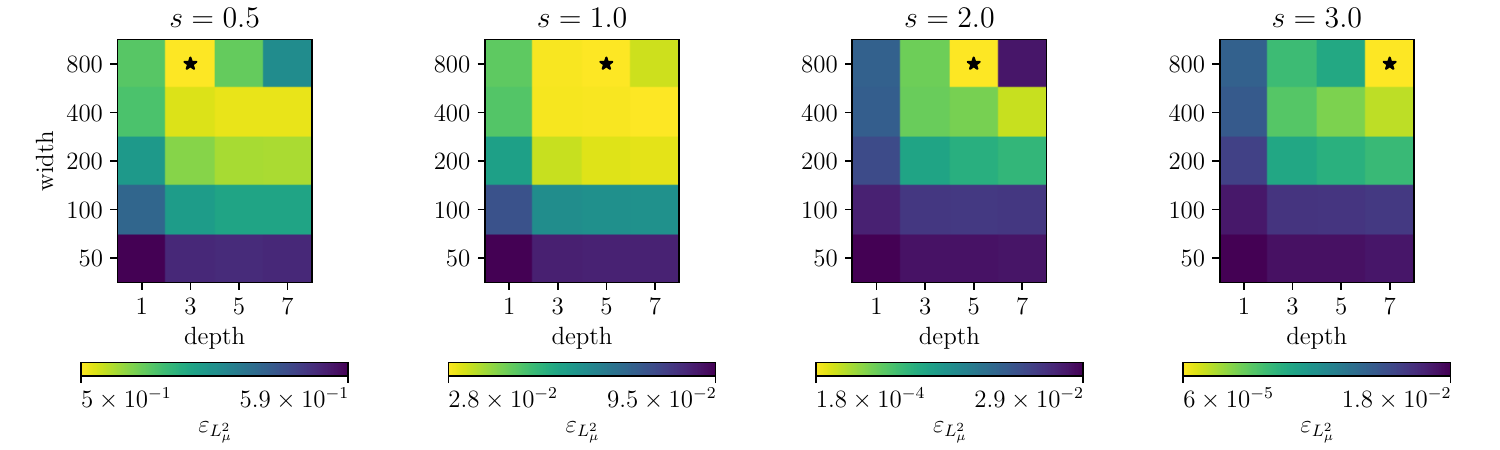}
    \caption{\small $\varepsilon_{L^2_\mu}(\HOneRBNO)$ for the diffusion problem \eqref{eq:elliptic} over network depth and width for different input smoothness $s$. Coefficient-space dimensions fixed at $d_{\rm in}=d_{\rm out}=200$, training set size $n=10^4$.}
    \label{fig:ops_hyperpars_rbdino}
\end{figure}

\begin{figure}
    \centering
    \includegraphics[width=\linewidth]{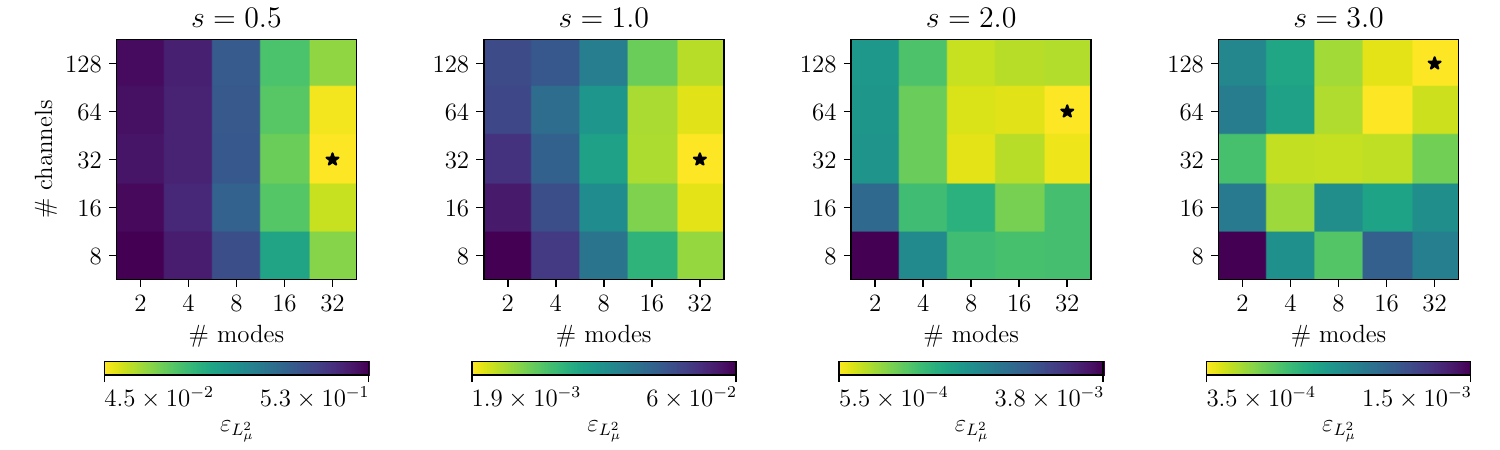}
    \caption{\small $\varepsilon_{L^2_\mu}(\FNO)$ for the diffusion problem \eqref{eq:elliptic} over the number of channels and Fourier modes for different input smoothness $s$. Network depth $L=4$, training set size $n=10^4$.}
    \label{fig:ops_hyperpars_fno}
\end{figure}

In terms of depth and width, the results for $\LTwoRBNO$ in \Cref{fig:ops_hyperpars_rbno} show more variation in which combination maximizes the accuracy.
Throughout all settings, neither very small nor very large networks are consistently superior.
This aligns with the intuition that small networks lack expressivity, while large networks may be challenging to optimize.
For $\HOneRBNO$, in contrast, \Cref{fig:ops_hyperpars_rbdino} shows a stronger correlation between network size (in terms of depth and width) and accuracy. This could be an indication that the more informative loss function allows to better leverage the increased expressivity of the architecture. For $\FNO$, we also observe an increase in accuracy as the network size, here controlled by the number of channels and Fourier modes, increases, as shown in \Cref{fig:ops_hyperpars_fno}.

\subsubsection{Sparse polynomial spaces} \label{sec:hyperpars_ip}

Recall that the polynomial spaces $\mathbb{P}_\Lambda$ in which we approximate the coefficients of the operator output are characterized by multi-index sets of the form \eqref{eq:ip_lambda}. While the parameter $\ell$ controls the size of the multi-index set, the parameters $a > 1$ and $b > 0$ allow varying the anisotropy of the polynomial space.

\begin{figure}
  \centering
  \includegraphics[width=\textwidth]{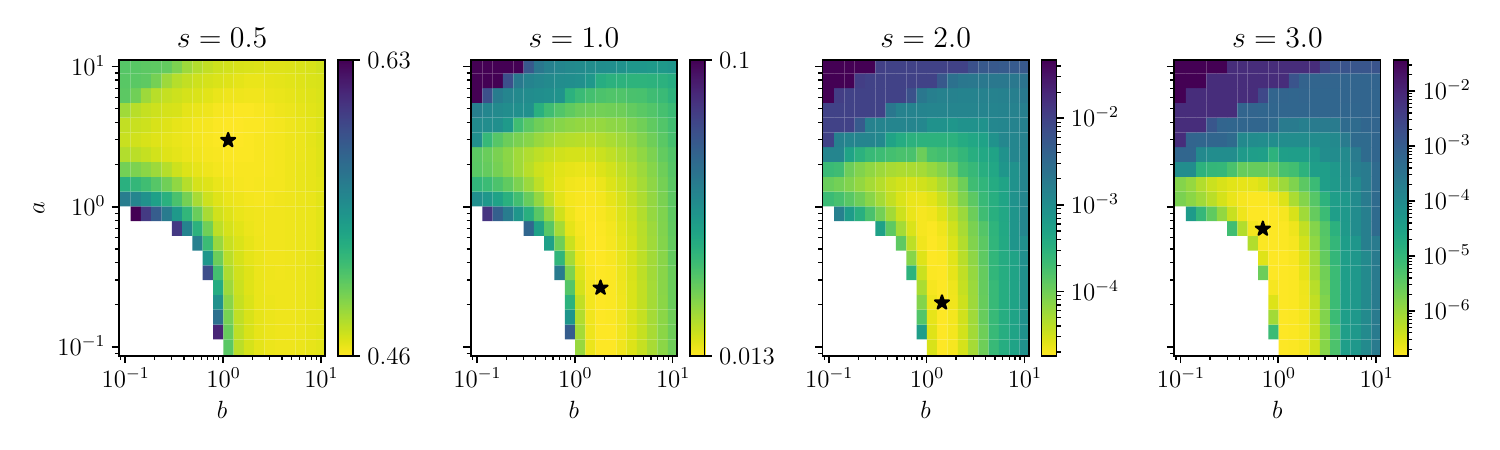}
  \caption{$\varepsilon_{L^2_\mu}(\Smolyak)$ for the diffusion problem \eqref{eq:elliptic} in dependence of the interpolation hyperparameters $a$ and $b$ as in \eqref{eq:ip_lambda} for different operator input smoothness $s$, with the minimum of the error landscape marked by $\star$. Fixed $d_{\rm in} = d_{\rm out} = 200$ and $n=10^4$.}
  \label{fig:p_ip_hyperpars}
\end{figure}

\Cref{fig:p_ip_hyperpars} shows how the surrogate error depends on these two parameters in the diffusion problem and for different smoothness $s$ of the operator input. In the observed parameter range, the error landscape exhibits a single minimum. The specific location of the minimum varies across different input smoothness $s$ and across test problems, as well as in dependence of the parameter amplitude and encoder and decoder ranks.

\subsubsection{Encoder-decoder setup}\label{ssec:enc_dec_setup}

\begin{figure}
    \centering
    \includegraphics[width=\textwidth]{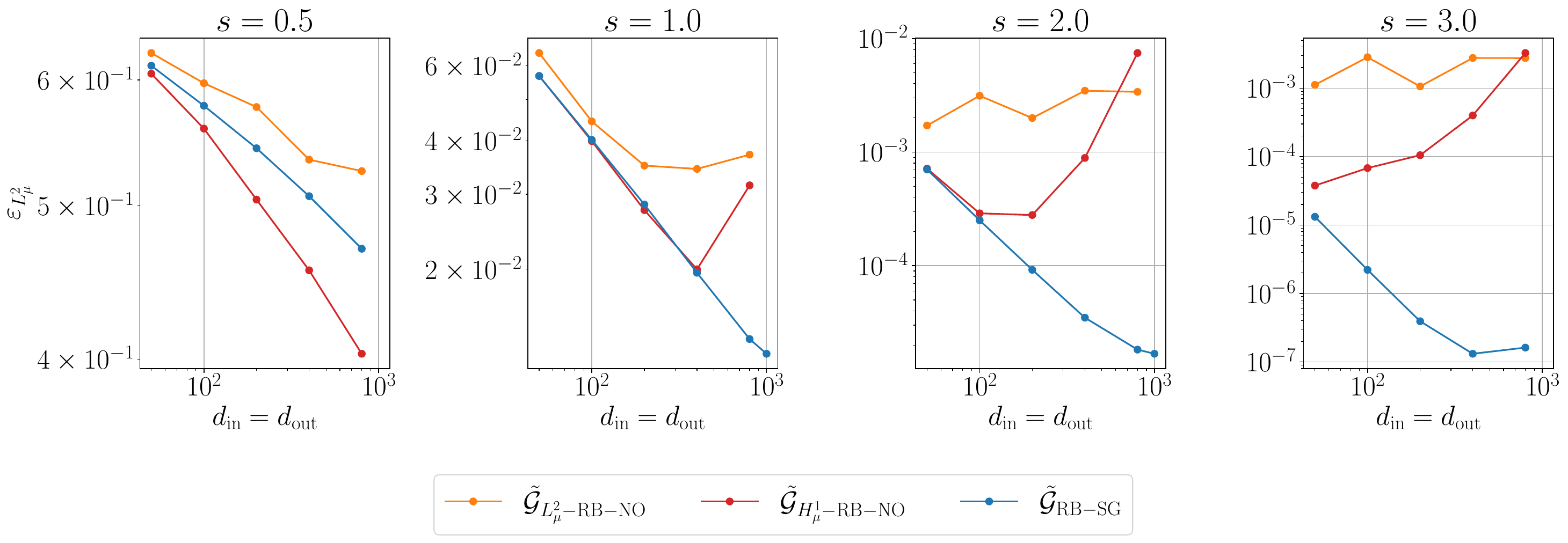}
    \caption{$\varepsilon_{L^2_\mu}$ for the diffusion problem \eqref{eq:elliptic} in dependence of encoder and decoder dimensions, $d_{\rm in}$ and $d_{\rm out}$, for various values of $s$ and $n=10^4$ training samples. $\LTwoRBNO$ uses networks of depth $7$ (for $s\in\{0.5,1.0\}$) or depth $9$ (for $s\in\{2.0, 3.0\}$) and width $200$, while $\HOneRBNO$ uses networks of depth $7$ and width $400$. $\Smolyak$ is constructed using hyperparameters $a=0.5$ and $b=1.2$.	These hyperparameters were selected to lie within regimes of low surrogate error as shown in \Cref{fig:ops_hyperpars_rbno,fig:ops_hyperpars_rbdino,fig:p_ip_hyperpars}.}
  \label{fig:dims}
\end{figure}

When setting up an encoder-decoder architecture, one has to choose the number of input and output dimensions, $d_{\rm in}$ and $d_{\rm out}$. We examine the impact of this choice on surrogate accuracy in \Cref{fig:dims}, where we study the convergence of $\varepsilon_{L^2_\mu}$ as $d_{\rm in}$ and $d_{\rm out}$ (simultaneously) increase up to $10^3$, which is the number of input dimensions used to generate the ground truth data (see \Cref{sec:parameter_field}). We observe that for small values of $s$, the error decreases as the number of dimensions $d = d_{\rm in} = d_{\rm out}$ increases. For larger $s$, however, there appears to be a threshold beyond which increasing $d$ no longer improves surrogate accuracy and even has a negative impact. This threshold can be observed for all three surrogate types, but appears earlier (both in terms of  $s$ as well as $d$) for the neural surrogates $\LTwoRBNO$ and $\HOneRBNO$ than for the polynomial surrogate $\Smolyak$.

\begin{figure}[htbp]
    \centering
    \begin{minipage}[t]{0.48\textwidth}
        \centering
        \includegraphics[width=\textwidth]{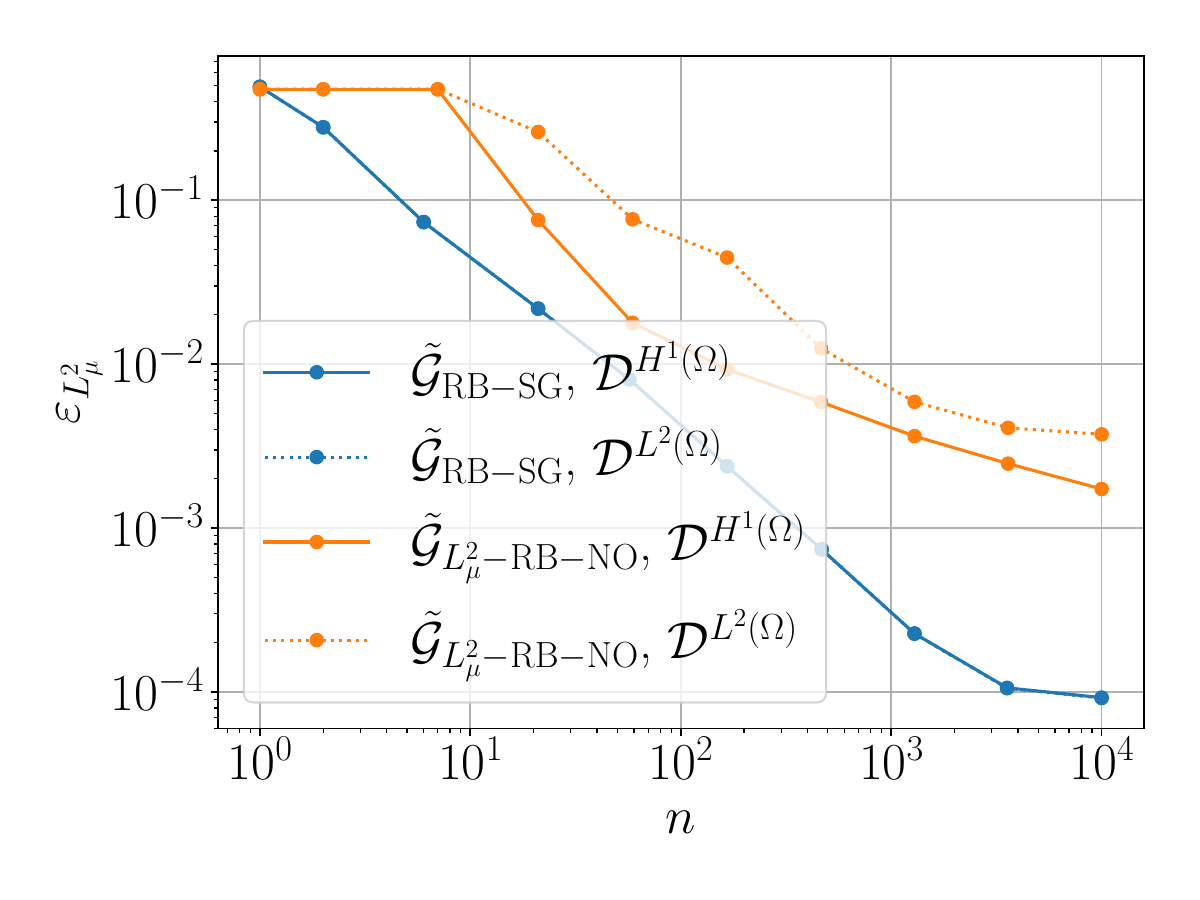}
        \caption{$\varepsilon_{L^2_\mu}(\tilde{\CG})$ in the diffusion problem \eqref{eq:elliptic} using $H^1(\Omega)$-orthogonal or $L^2(\Omega)$-orthogonal basis functions to construct the decoder $\CD$. Fixed $d_{\rm in} = d_{\rm out} = 200$ and $s=2$. $\Smolyak$ is constructed using hyperparameters $a=b=1$ and $\LTwoRBNO$ using depth $9$ and width $100$.}
        \label{fig:input_orthogonalization}
    \end{minipage}
    \hfill
    \begin{minipage}[t]{0.48\textwidth}
        \centering
   	    \includegraphics[width=\textwidth]{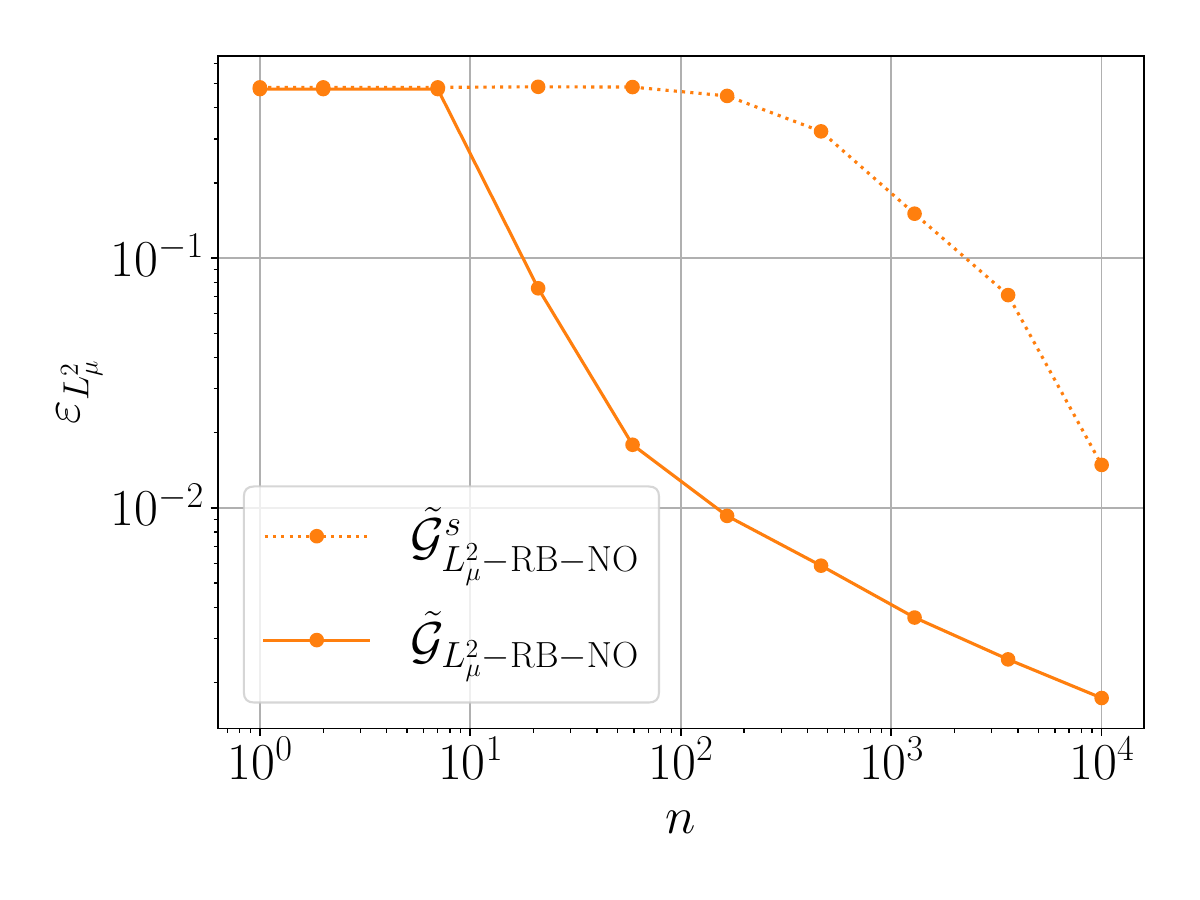}
        \caption{$\varepsilon_{L^2_\mu}$ of $\LTwoRBNO$ and $\LTwoRBNO^s$ as in \eqref{eq:scaling_rbno} in the diffusion problem \eqref{eq:elliptic}. Fixed $d_{\rm in} = d_{\rm out} = 200$, $s=2$ and network depth $9$ and width $100$.}
        \label{fig:scaling_rbno}
    \end{minipage}
\end{figure}

We next consider the orthogonalization of the decoder basis.
Since we measure surrogate accuracy in the $L^2(L^2(\Omega),\mu;H^1(\Omega))$ norm, where $\Omega$ denotes the physical domain of the PDE, it is natural to choose a decoder basis that is orthonormal with respect to the $H^1(\Omega)$ inner product.
We study the robustness of surrogate accuracy with respect to the choice of this inner product in \Cref{fig:input_orthogonalization}.
There, we compare surrogate accuracy of $\LTwoRBNO$ and $\Smolyak$ constructed with the default $H^1(\Omega)$-orthonormal decoder basis, to versions constructed using an $L^2(\Omega)$-orthonormal basis.  We see our intuition confirmed for $\LTwoRBNO$, which is able to achieve better accuracy when using an $H^1(\Omega)$-orthonormal decoder basis.
Since the (linear) interpolant is invariant with respect to a linear transformation of the interpolated function, $\Smolyak$ is invariant to the choice of basis, as long as the span of the basis vectors is the same; especially for a small truncation number $d_{\rm out}$ in the output representation, the choice of the basis (and thus the underlying linear space) becomes more critical however.

Finally, we examine the role of input normalization, which is well known to improve neural network training speed and stability \cite{ioffe2015batchnormalizationacceleratingdeep, santurkar2019doesbatchnormalizationhelp}. Motivated by this, we consider a deterministic rescaling of the encoder coefficients $\CE(x) \in \bigtimes_{i \in \N} [-\lambda_i^s, \lambda_i^s]$ to $[-1,1]^\N$ by absorbing the factors $(\lambda_i^s)_{i\in \N}$ into the encoder basis. Replacing $(\psi_i)_{i \in \N}$ with $(\lambda_i^{-s} \psi_i)_{i \in \N}$, we obtain a rescaled encoder
\begin{equation*}
    \CE^s(x) := (\langle x-m_\CX, \lambda_i^{-s} \psi_i\rangle_{\CX})_{i\in\N} \in  [-1,1]^\N , \quad \text{ for } x \in K^s.
\end{equation*}
We then approximate $g^s = \CD^{-1} \circ \CG \circ (\CE^{s})^{-1}  \colon [-1,1]^\N \to \ell^2(\N)$ with $\tilde{g}^s$ and set
\begin{equation}\label{eq:scaling_rbno}
\tilde{\CG}^s := \CD \circ \tilde{g}^s \circ \CE^s.
\end{equation}
However, the results in \Cref{fig:scaling_rbno} show that training neural networks to approximate $g^s$ leads to significantly lower surrogate accuracy than training to approximate $g$. A possible explanation is that the input scales encode their relative importance and rescaling them to a uniform range may suppress this information and slow convergence \cite{lecun2002efficient}.
We point out that for polynomial surrogates, approximating $g$ or $g^s$ is purely an implementation choice (namely, the interval to which interpolation nodes are scaled) and does not impact accuracy.

\subsubsection{TT surrogates}\label{ssec:tt_setup}

As shown in \Cref{sec:TT}, the key factor in size and evaluation cost for TT surrogates are the ranks.
We study suitable choices for several aspects of the ALS cross algorithm in order to minimize them.

\paragraph{Starting ranks and initialization.}

The ALS cross algorithm starts from a, usually randomly initialized, tensor with user-chosen starting ranks.

There are multiple strategies available:
\begin{itemize}
    \item Start with small initial ranks and adaptively grow ranks using the AMEn \cite{Dolgov_2014} algorithm.
    In order to achieve higher accuracy, the iteration count is increased.

    \item Initialize with fixed, a priori chosen ranks and run only few iterations.
    This yields a tensor of specific size, determined by the starting ranks.

    \item Derive an estimate of the solution ranks from the TT ranks of the input parameter and run only a few iterations.
      The accuracy is increased by supplying a more accurate TT representation of the input parameter.
      If this TT representation is available with high accuracy, TT rounding to a given tolerance can be used before supplying it to the ALS cross algorithm.
\end{itemize}

\begin{figure}[H]
    \centering
    \includegraphics[width=\textwidth]{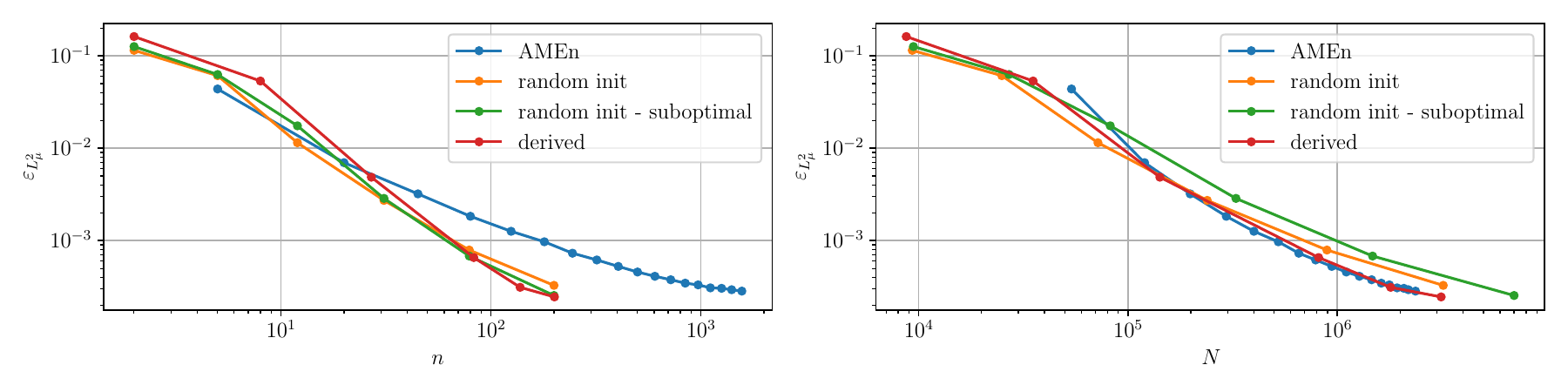}
    \caption{
    $\varepsilon_{L^2_\mu}(\TT)$ for the elliptic PDE \eqref{eq:elliptic} over number of PDE solves $n$ and TT surrogate size $N$ for
    $d_{\rm in} = 100$ and $s=2.0$.
    }
    \label{fig:als_init_strat}
\end{figure}

In \Cref{fig:als_init_strat}, we compare the three approaches.
In terms of error over PDE evaluations, the AMEn iteration is clearly less efficient, while the other two approaches are quite similar.
Looking at the error over TT size $N$, AMEn and derived ranks yield slightly better results at higher accuracy.
We also include results for the random initialization approach with a smaller truncation tolerance.
This suboptimal choice leads to an increased TT size with little benefit to accuracy, highlighting the need for a careful choice of this parameter.

\paragraph{ALS iteration.}

\begin{figure}[H]
    \centering
    \includegraphics[width=\textwidth]{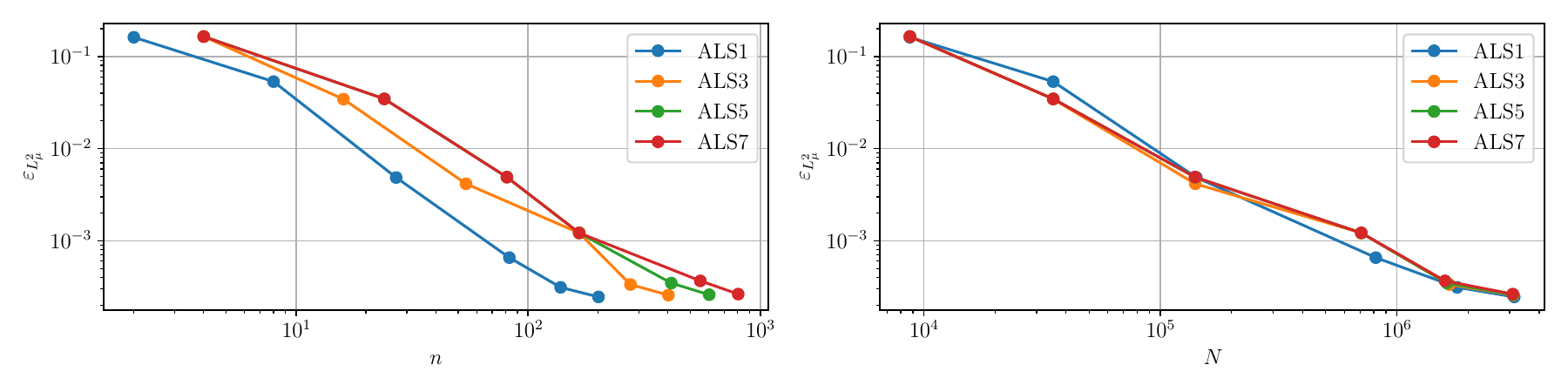}
    \caption{
      $\varepsilon_{L^2_\mu}(\TT)$
      for the elliptic PDE \eqref{eq:elliptic}
      over number of PDE solves $n$ and TT surrogate size $N$ for
    $d_{\rm in} = 100$, $\theta=1.0$ and $s=2.0$.
    ALSx performs x end-to-end iterations over the TT.
    }
    \label{fig:als_iteration_count}
\end{figure}

The ALS cross uses alternating optimization over the cores of the tensor.
Each forward iteration requires new evaluations of the PDE operator in addition to the rest of the computation on the reduced systems.
Thus, we need to strike a balance between converging the ALS iteration, increasing accuracy, and
cost of computing the surrogate.
In practice, we observe that very few iterations are necessary. In fact, even just one forward
iteration yields good results for the elliptic PDE \eqref{eq:elliptic}.
\Cref{fig:als_iteration_count} shows how increasing the number of ALS iterations increases the number of PDE solves $n$, while accuracy and resulting TT size see no meaningful improvements.

\paragraph{Polynomial degrees.}

The ALS cross algorithm requires an a priori choice of the number of collocation points $\nu_k +1$ along each parametric dimension.
This is equivalent to choosing the polynomial degree along that dimension; while higher values generally lead to higher accuracy, they can also introduce instabilities due to high-order polynomial interpolation.
Furthermore, increasing $\nu_k$ increases the size of the TT cores and thus leads to larger, less performant TT representations.

In practice, the total error of the surrogate comprises several other error terms.
We therefore choose $\bsnu$ minimal such that the interpolation error is of the same order as the other error terms.
In the absence of a good heuristic, we need to determine $\bsnu$ empirically.
Experience suggests that single-digit values typically suffice.

To take advantage of the decay properties of the input coefficients, we can use a decaying $\bsnu$ such as
\begin{equation}\label{eq:points_aniso}
    \nu_k = \left\lceil \frac{\nu_{max}}{\log_2(k+1)} \right\rceil - 1, \qquad k \in \N
\end{equation}
for some choice of $\nu_{max} >1$ and set the number of input dimensions to at most
$d_{\rm in} = \max \{k \in \N \ \vert \ \nu_k > 0\}$.

\begin{figure}[H]
    \centering
    \includegraphics[width=\textwidth]{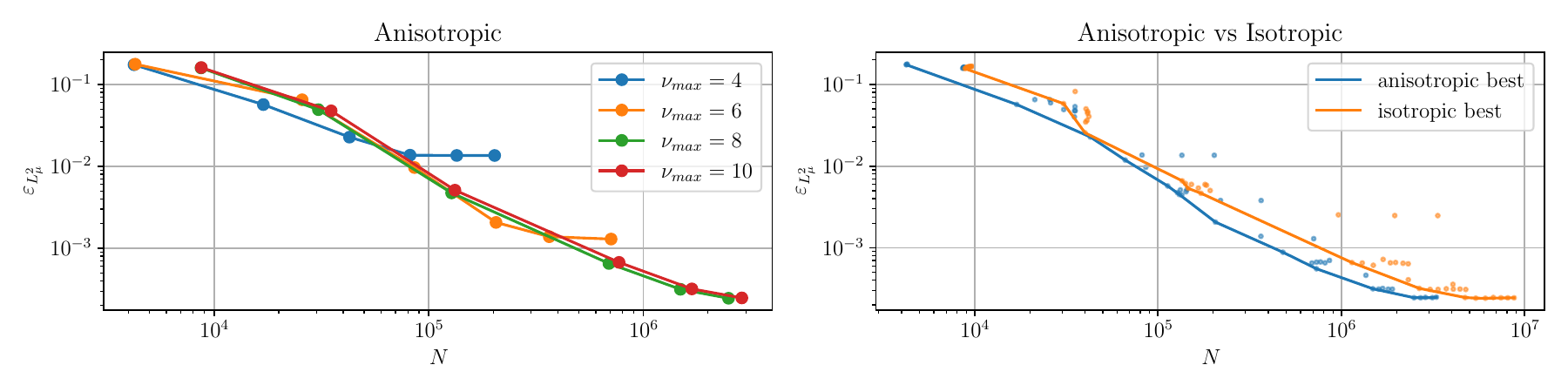}
    \caption{
    $\varepsilon_{L^2_\mu}(\TT)$ for the elliptic PDE \eqref{eq:elliptic} over TT surrogate size $N$ for $d_{\rm in} = 100$ and $s=2.0$.}
    \label{fig:als_iso_aniso}
\end{figure}

In \Cref{fig:als_iso_aniso} we show the impact of the number of collocation points on the accuracy/size tradeoff for anisotropic $\bsnu$ as in \eqref{eq:points_aniso} on the left.
We observe the polynomial degree becoming a bottleneck if it is not chosen high enough for the desired accuracy.
However, as long as it is sufficient for the desired accuracy, increasing the polynomial degree has little impact.
Thus, it is usually best to choose a moderate value that is high enough.
On the right, we illustrate the advantage of the anisotropic choice. Compared to isotropic $\nu_k = \nu_{\max}$ for some $\nu_{\max} \in \N$, the choice \eqref{eq:points_aniso} improves the accuracy/size tradeoff across the board.

\subsection{Performance} \label{sec:exp_convergence}

We now compare the performance of operator surrogates on the two test problems introduced in \Cref{sec:exp_probs}. Since surrogates are supposed to be not only accurate but also computationally cheap approximations of the true operator, our empirical study focuses on their cost–accuracy trade-off. As described in \Cref{sec:measuring}, we quantify surrogate accuracy by Monte Carlo approximations of the $L^2_\mu$- and $H^1_\mu$-errors.
Surrogate cost is decomposed into three components, which we measure separately. First, we use the number of training samples as a proxy for data generation cost. Second, we record the setup cost, defined as training time for neural operators and initialization cost for polynomial operator surrogates. Third, we measure the evaluation time, which reflects the computational cost of applying the surrogate in downstream tasks.

The hyperparameters examined in the previous section affect not only accuracy but also setup and evaluation costs, and different configurations may yield different cost–accuracy trade-offs. To enable a comprehensive comparison, we analyze the trade-off across ensembles of surrogates generated by varying hyperparameter settings. Based on the findings in \Cref{sec:exp_hyperpars}, we restrict the hyperparameter ranges only where limited by model constraints, deteriorating performance with increasing complexity, or prohibitive training cost. The resulting configurations are summarized in \Cref{table:parameters}.
For each experiment and surrogate type, we display the convex hull of outcomes obtained across hyperparameter choices and highlight the corresponding Pareto frontier representing optimal cost–accuracy trade-offs.

\begin{table}[!h]
\centering
\begin{tabular}{|c|l|}
\hline
\textbf{Model} & \textbf{Parameters} \\
\hline
$\LTwoRBNO$ &
\begin{tabular}{ll}
$d_{\rm in}=d_{\rm out}$ & 50, 100, 200, 400, 800 \\
width & 200, 400, 800 \\
depth & 3, 5, 7 \\
\end{tabular} \\ \hline
$\HOneRBNO$ &
\begin{tabular}{ll}
$d_{\rm in}=d_{\rm out}$ & 50, 100, 200, 400, 800 \\
width & 200, 400, 800 \\
depth & 3, 5, 7 \\
\end{tabular} \\ \hline
$\FNO$ &
\begin{tabular}{ll}
modes & 2, 4, 8, 16, 32 \\
width & 8, 16, 32, 64 \\
\end{tabular} \\ \hline
$\Smolyak$ &
\begin{tabular}{ll}
$d_{\rm in}=d_{\rm out}$ & 50, 100, 200, 400, 800 \\
$a$ & 0.2, 0.5, 1.2, 3.0 \\
$b$ & 0.2, 0.5, 1.2, 3.0 \\
\end{tabular} \\ \hline
$\TT$ &
\begin{tabular}{ll}
$d_{\rm in}$ & 50, 100, 200 ($s \geq 1)$ \\
& 10, 15, 20, 25 ($s = 0.5)$ \\
\end{tabular}
\\
\hline
\end{tabular}
\caption{Architectural parameters used throughout the experiments in \Cref{sec:exp_convergence}.}
\label{table:parameters}
\end{table}

\subsubsection{$L^2_\mu$-error vs number of training data $n$}\label{sec:exp_error_training_data}

\begin{figure}
  \centering
  \begin{minipage}[t]{\textwidth}
  \includegraphics[width=\textwidth]{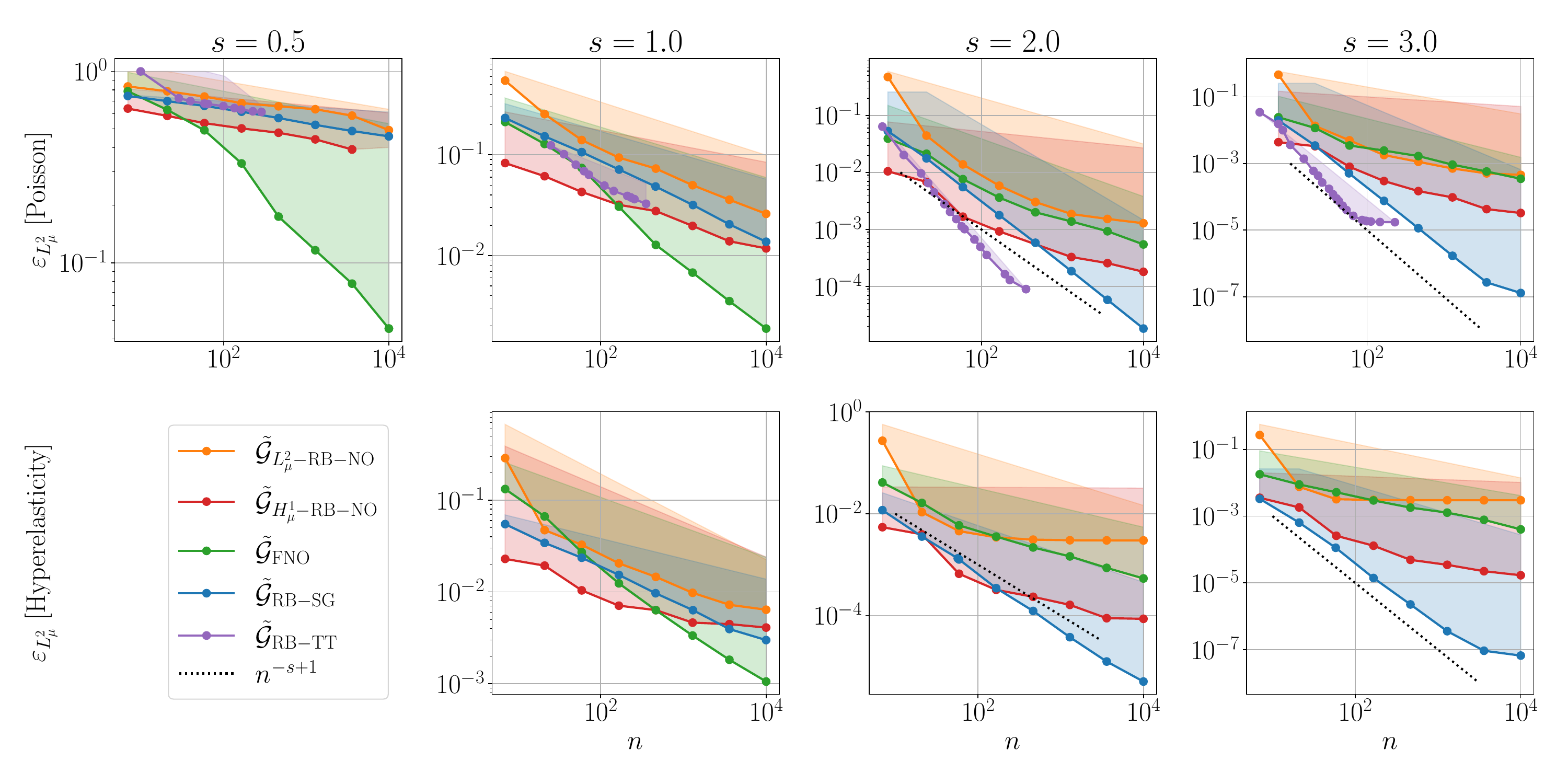}
    \caption{$\varepsilon_{L^2_\mu}$-convergence over number of training data $n$ for different $s$.}
  \label{fig:cato_solves}
  \end{minipage} \vspace{1.5cm}\\
  \begin{minipage}[t]{\textwidth}
    \centering
    \includegraphics[width=1\linewidth]{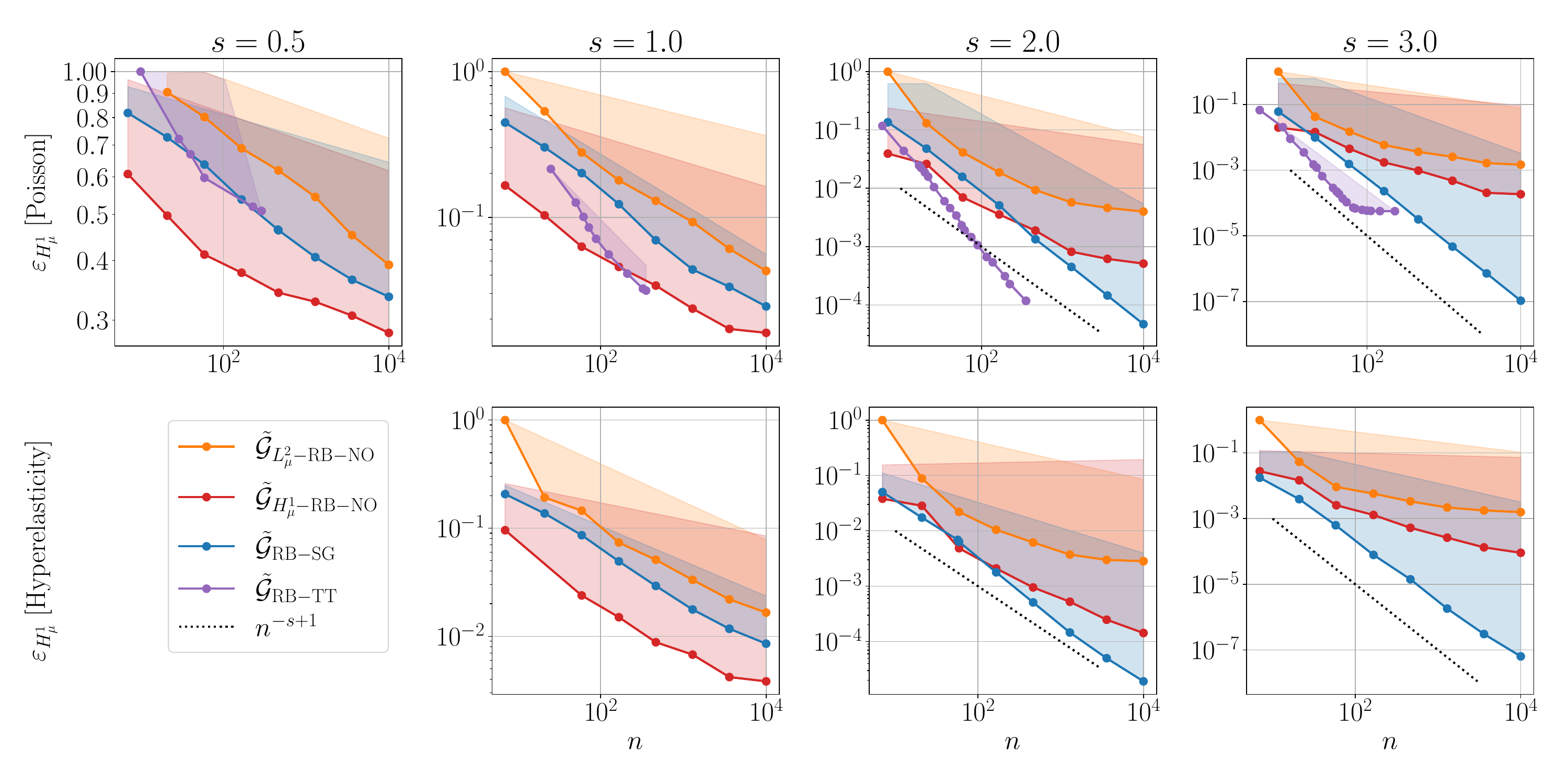}
    \caption{$\varepsilon_{H^1_\mu}$-convergence over number of training data points $n$ for different $s$.}
    \label{fig:jac_error_solves}
  \end{minipage} \hfill
\end{figure}

Supervised surrogate architectures, as considered in this work, require operator evaluations for training or setup. Since these evaluations are costly in real-world applications, identifying data-efficient methods is critical. We therefore study the trade-off between surrogate error and the number of operator evaluations used for training or setup in \Cref{fig:cato_solves}. For each surrogate, we report the range of outcomes achieved with the hyperparameter configurations in \Cref{table:parameters} and identify the Pareto-optimal trade-offs for varying input smoothness $s$.

We observe that convergence improves as input smoothness increases across all surrogate types. The polynomial surrogates $\Smolyak$ and $\TT$, however, exploit input smoothness more effectively, achieving better convergence rates compared to neural operator surrogates for $s \in \{2.0,3.0\}$. In this regime, the convergence of $\Smolyak$ (roughly) follows the theoretical rates predicted by \Cref{thm:ops:interpolation}. We observe convergence also in the regime not covered by \Cref{thm:ops:interpolation} since the test problem uses a large but finite number of input dimensions. The convergence rates deteriorate as $s$ decreases, as (higher order) derivatives of the coefficient map tend to become larger. Both methods have similar convergence rates, with $\TT$ seemingly having a better constant. However, $\TT$ showcases a plateauing error earlier than $\Smolyak$, especially apparent for $s=3$. While the root cause of this behavior is still under investigation, there seems to be some numerical instability in the TT algorithms, preventing further convergence. For similar reasons $\TT$ requires special attention for $s=0.5$, diverging for $r_i > 25$.

Comparing the derivative informed method $\HOneRBNO$ to $\LTwoRBNO$, which uses the same underlying network architecture, efficiency is improved. However, it should be noted that $\HOneRBNO$, contrary to other methods, requires Jacobian information of $\tilde{g}$ for training. This is not a limiting factor when the acquisition of training data is the bottleneck and derivative samples can be obtained at little extra cost (cf.\ \Cref{sec:derivative_sample_gerneation}). However, the training process itself is significantly more expensive for $\HOneRBNO$.

For $s \in \{0.5, 1.0\}$, $\FNO$ achieves better convergence rates than the other surrogate types. On the smoother input fields with weaker high-frequency components, this advantage disappears.

Lastly, these observations do not change significantly for the nonlinear hyperelasticity problem.

\subsubsection{$H^1_\mu$-error vs number of training data $n$}

We now consider $\varepsilon_{H^1_\mu}$ as in \eqref{eq:H1mu}, i.e., the error of the derivative of the surrogate with respect to the parametric input. For neural surrogates, the Jacobian of $\tilde{g}$ is computed via automatic differentiation. For polynomial surrogates, it is straightforward to implement the analytic expression for the Jacobian of $\tilde{g}$, which is faster to evaluate than using automatic differentiation.
Since $\TT$ computes the coefficients of the output with regard to the finite element discretization directly, we use $\nabla (\CD^\dagger \TT)$ for these comparisons.
$\FNO$ is not included in the following results, as this architecture does not naturally produce derivatives of the coefficient map. Note, however, that derivative information of the operator can be included in FNO training for improved accuracy \cite{yao2025derivative}.

The results displayed in \Cref{fig:jac_error_solves} indicate relative strengths and weaknesses of the reduced-basis surrogate architectures that are consistent with those observed in the previous section. In all cases, $\HOneRBNO$ outperforms $\LTwoRBNO$. The polynomial surrogates achieve competitive convergence rates, in particular for smooth inputs with $s \in \{2,3\}$. Neural operators, however, exhibit a smaller convergence constant for rougher data with $s \in \{0.5,1\}$. Finally, note that the polynomial surrogates do not require derivative information during training or setup.

\subsubsection{$L^2_\mu$-error vs evaluation time $t_E$}\label{sec:exp_error_eval_cost}

For operator surrogates to serve their purpose, low online cost to evaluate at new input values is crucial.
To assess the methods with respect to this criterion, we report the approximation error as a function of the surrogate evaluation time.
Note that runtime measurements compare not only the methods themselves, but also their software implementations and the hardware used. Although care has been taken to ensure efficient implementations, further optimization may be possible. All surrogates are implemented in either PyTorch or JAX, and the runtime data shown in the following was obtained on a Nvidia H200 GPU. Different hardware might showcase different scaling behavior.

\begin{figure}[h]
  \centering
  \includegraphics[width=\textwidth]{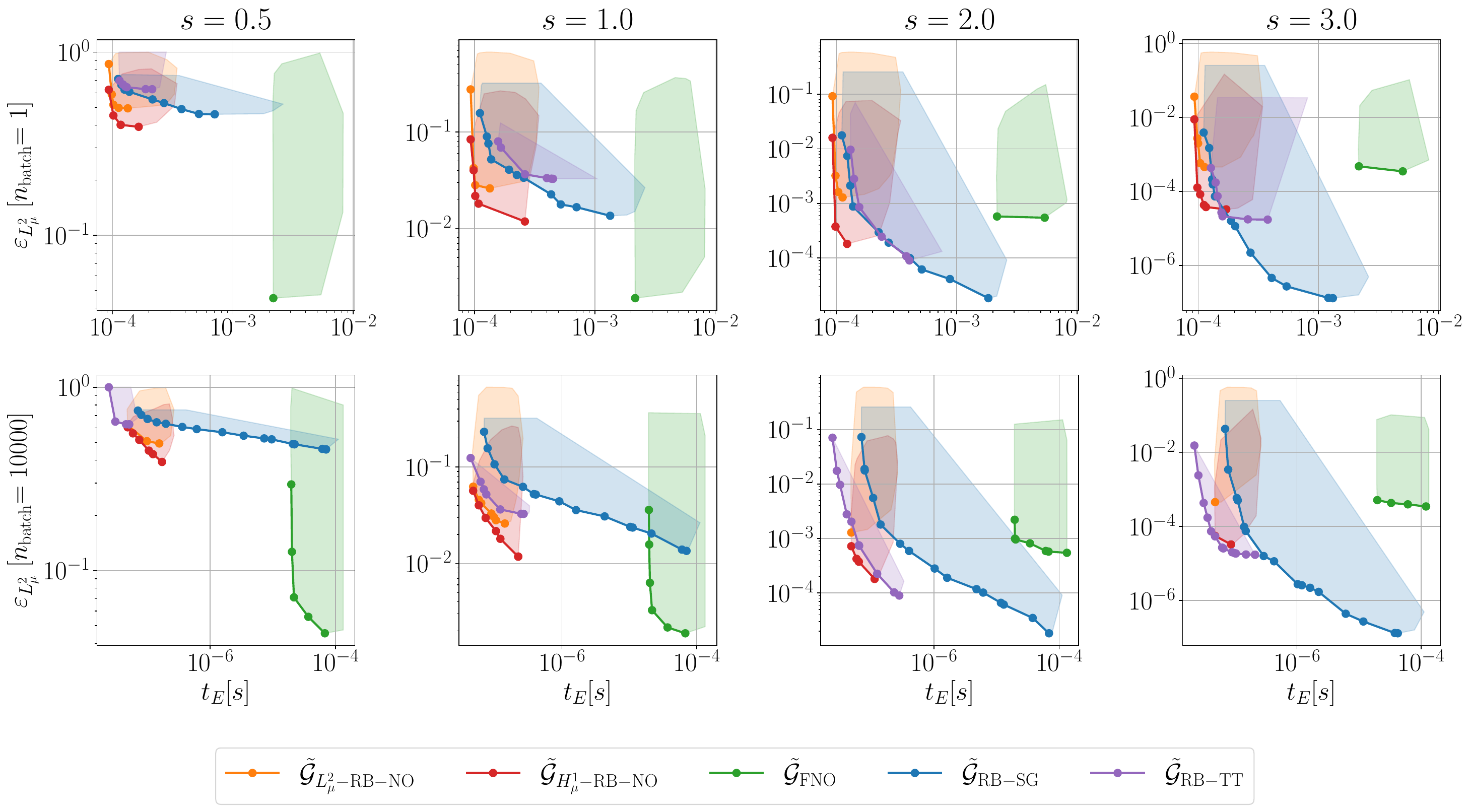}
  \caption{$\varepsilon_{L^2_\mu}$-error versus surrogate evaluation time $t_E$ per sample for the diffusion problem \eqref{eq:elliptic}. The top row shows results obtained when evaluating a single sample, while the bottom row shows evaluation time per sample when evaluating the surrogates on a batch of size $n_{\rm batch} = 10^4$, for different $s$.}
  \label{fig:cato_eval}
\end{figure}

Results are shown in \Cref{fig:cato_eval}, aggregated over the range of training data volumes used in the previous subsections. For polynomial surrogates, data volume affects both accuracy and computational cost, whereas for neural operator surrogates, cost is determined by the architecture and is thus independent of data volume. Consequently, the result ranges of neural operator surrogates exhibit greater variation in accuracy than in cost, while for polynomial surrogates, an increase in accuracy typically correlates with an increase in cost.

We observe that for lower accuracy requirements, all surrogates except $\FNO$ are competitive. When higher accuracy is needed, one must resort to a more expensive surrogate ($\FNO$ for rougher inputs, $\Smolyak$ for smoother inputs).

\begin{figure}
    \centering
    \includegraphics[width=.9\linewidth]{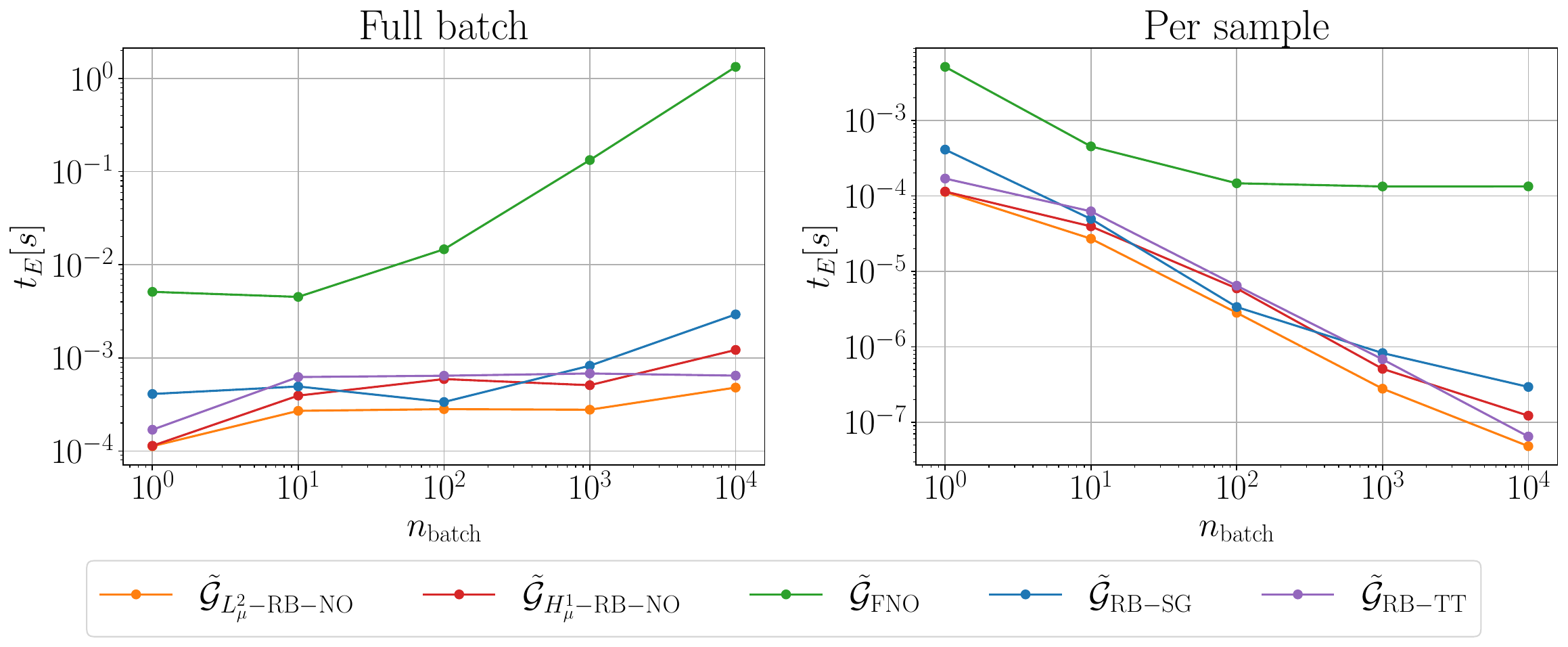}
    \caption{Evaluation cost over batch size for $s=2$ and $\varepsilon_{L^2_\mu} \sim 10^{-3}$ for the diffusion problem \eqref{eq:elliptic}.}
    \label{fig:batch_size_comparison}
\end{figure}

We also want to point out that the different surrogates scale differently with the batch size.
Some methods and their respective implementations have higher fixed runtime costs, making them more attractive for large batch sizes. We show this scaling for a fixed choice of $s=2$ and at $\varepsilon_{L^2_\mu} \sim 10^{-3}$ in \Cref{fig:batch_size_comparison}. The reduced-basis surrogates $\LTwoRBNO$, $\HOneRBNO$, $\Smolyak$ and $\TT$ scale well with batch size $n_{\rm batch}$, yielding lower per-sample evaluation time $t_E$ as the batch size increases. In contrast, $\FNO$ quickly saturates and offers only constant per-sample evaluation time $t_E$. This is likely because intermediate results have a large memory footprint, allowing only a limited number of samples to be processed in parallel.

\begin{figure}[h]
  \centering
  \includegraphics[width=\textwidth]{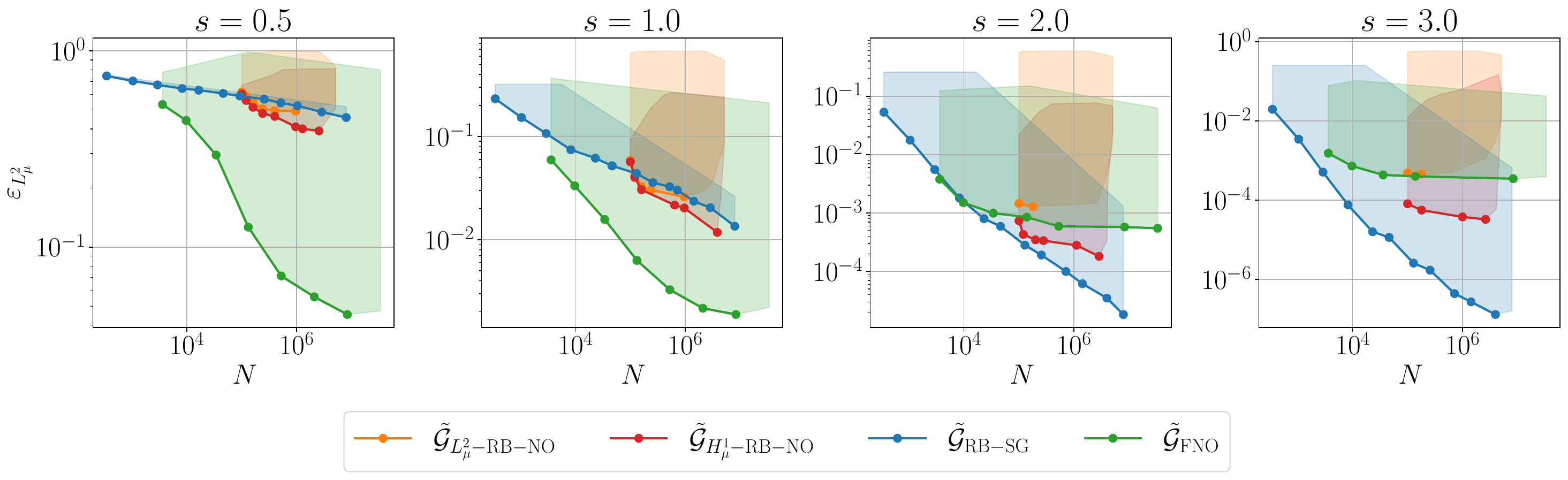}
    \caption{$\varepsilon_{L^2_\mu}$-convergence over parametric surrogate complexity $N$ for different $s$  for the diffusion problem \eqref{eq:elliptic}.}
  \label{fig:cato_parametric}
\end{figure}

As surrogate evaluation time may depend on implementation and hardware specifics, we further show accuracy versus parametric complexity of the surrogates in \Cref{fig:cato_parametric}. We understand parametric complexity as the number $N$ of free parameters used in a surrogate. For $\Smolyak$, we have $N = d_{\rm out} n$, where $n$ is the number of interpolation points used. For neural operator surrogates, $N$ is given by the total number of weights and biases. For the TT surrogate, $N$ is the number of variables in the cores.

\subsubsection{$L^2_\mu$-error vs training time $t_T$}

Another consideration is the offline cost, i.e., the one-time setup cost of training the network or running other setup computations. When measuring surrogate setup time, again, software implementation as well as hardware properties may impact the results. For neural operator surrogates, setup cost is dominated by training, which we perform on an Nvidia A100 GPU for the results shown here. Setting up sparse-grid surrogates is comparatively straightforward, as it involves only setting up necessary data structures as well as a warm-up call to trigger just-in-time compilation of the JAX backend. For the TT surrogate, setup is dominated by the ALS cross algorithm, which is implemented in NumPy and thus runs on CPU.

\begin{figure}[h]
  \centering
  \includegraphics[width=\textwidth]{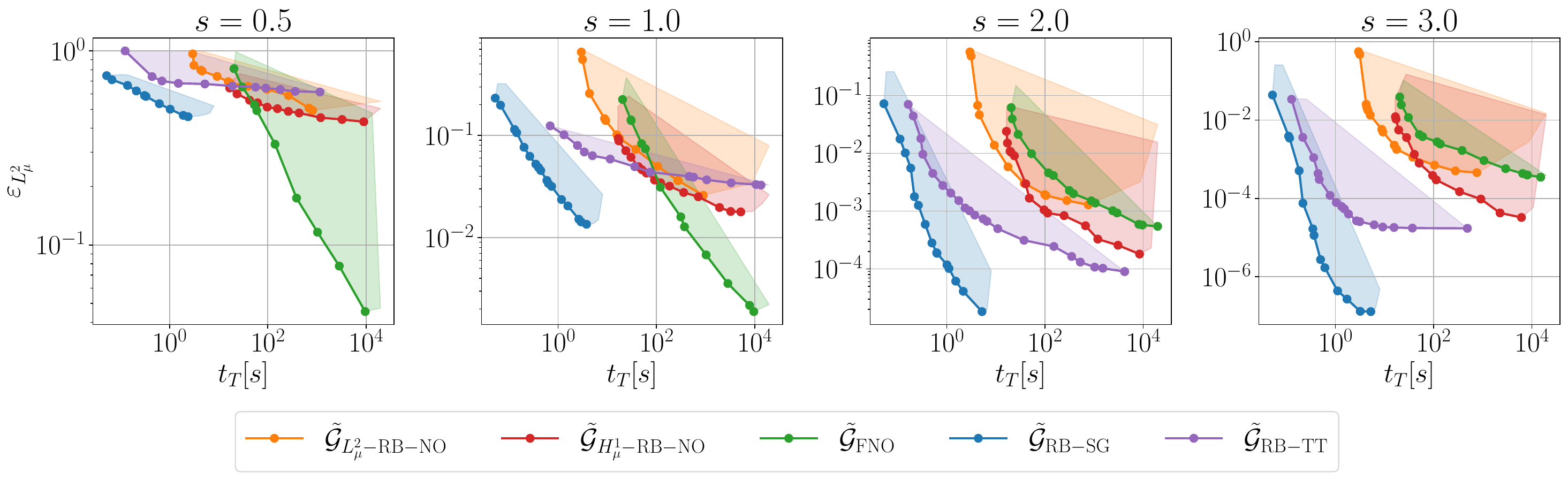}
  \caption{$\varepsilon_{L^2_\mu}$-convergence over training/setup time $t_T$ for different $s$ for the diffusion problem \eqref{eq:elliptic}.}
  \label{fig:cato_train}
\end{figure}

Results are shown in \Cref{fig:cato_train}.
Unsurprisingly, neural surrogates have much higher setup time since they require expensive training. Among them, $\LTwoRBNO$ is cheaper due to a simpler loss function, while $\FNO$ and $\HOneRBNO$ are roughly on par. The sparse-grid surrogate $\Smolyak$ requires only some data structures to be set up and is therefore very fast here. The tensor-train surrogate $\TT$ is somewhere in between, being quite fast at low to medium accuracy, but not scaling well to higher accuracy. The high ranks required to achieve high accuracy can make the ALS cross algorithm computationally expensive. Porting the code to run on GPUs might yield a significant speedup.

We emphasize that these results do not include the time to generate the training data, which is sometimes subsumed into the training cost, as for example in \citep{de2022cost}.

\section{Conclusions}

Overall, perhaps unsurprisingly, we see different surrogate methods showing their strengths in different areas. There is no outright ``best'' surrogate; instead, the right tool needs to be chosen for the application at hand. Furthermore, no method is guaranteed to work right out of the box. As highlighted in \Cref{sec:exp_hyperpars} and the ensemble plots of \Cref{sec:exp_convergence}, hyperparameter tuning is crucial to achieving good results.

One standout observation is how polynomial surrogates perform significantly better for smooth inputs than neural operators, both in terms of evaluation time and required amount of training data. For such problems, these methods seem clearly preferable. Among them, the TT surrogate is slightly more efficient with regard to the required samples. The sparse-grid surrogate, on the other hand, has much lower offline cost and offers striking simplicity, requiring only a method to evaluate $\tilde g$ at specific values, seems to be more stable, reach higher accuracy, and come with guaranteed convergence guarantees and a priori error estimates. However, the performance is to some extent reversed as the input data become rougher; in this case, the neural surrogates, and especially FNO, tend to have superior accuracy.

For the neural operator, we see that the basic $L^2_\mu$-RBNO struggles to compete, while the derivative-informed $H^1_\mu$-RBNO achieves better accuracy throughout. The main caveat is the availability of the derivative data in the first place, but if available at a reasonable cost, it should definitely be used.

The FNO responds quite differently to the input smoothness $s$ compared to the other methods. It performs exceptionally well for rough input data, where the other methods struggle to produce usable results. However, its convergence rates do not improve as quickly with increasing smoothness as those of polynomial operator surrogates, and for small $s$ its accuracy remains comparably low regardless of hyperparameter/architecture choices and amount of training data. In addition, the training and evaluation costs tend to be fairly high as a result of the complex architecture.

\bibliography{main}
\bibliographystyle{unsrtnat}

\end{document}